\documentclass{IEEEtran}

\usepackage{amsmath}
\usepackage{amssymb}
\usepackage{amsthm}
\ifCLASSINFOpdf \else \fi

\usepackage[dvips]{graphicx}
\usepackage{algorithm}
\usepackage{algorithmic}
\usepackage{slashbox}
\usepackage{subfigure}

\theoremstyle{definition}

\theoremstyle{Remark}

\usepackage{color}
\usepackage[normalem]{ulem}

\usepackage[numbers,sort&compress]{natbib}
\usepackage[table]{xcolor}
\usepackage{multirow}
\usepackage{makecell}
\usepackage{setspace}
\usepackage{stfloats}
\usepackage{supertabular}
\usepackage{verbatim}
\usepackage{threeparttable}
\usepackage{array}
\newcolumntype{L}[1]{>{\raggedright\let\newline\\\arraybackslash\hspace{0pt}}m{#1}}
\newcolumntype{C}[1]{>{\centering\let\newline\\\arraybackslash\hspace{0pt}}m{#1}}
\newcolumntype{R}[1]{>{\raggedleft\let\newline\\\arraybackslash\hspace{0pt}}m{#1}}
\usepackage{color}
\usepackage{xcolor}
\usepackage[mathscr]{eucal}
\usepackage{cases}
\usepackage{fancyvrb}

\UseRawInputEncoding

\begin{document}
\title{UAV Assisted Data Collection for Internet of Things: A Survey}

\author{Zhiqing Wei,~\IEEEmembership{Member,~IEEE,}~Mingyue Zhu,~Ning Zhang,~\IEEEmembership{Senior Member,~IEEE,}~Lin Wang,\\
	~Yingying Zou,~Zeyang Meng,~Huici Wu,~\IEEEmembership{Member,~IEEE,}~and~Zhiyong Feng,~\IEEEmembership{Senior Member,~IEEE}

\thanks{

Zhiqing Wei, Mingyue Zhu, Lin Wang, Yingying Zou, Zeyang Meng, and Zhiyong Feng are with the Key Laboratory of Universal Wireless Communications, Ministry of Education, School of Information and Communication Engineering, Beijing University of Posts and Telecommunications, Beijing 100876, China (email: \{weizhiqing; mingyue\_zhu; wlwl; zouyingying; mengzeyang; fengzy \}@bupt.edu.cn).

Ning Zhang is with the Department of Electrical and Computer Engineering, 
University of Windsor, Windsor, ON, N9B 3P4, Canada. (e-mail: ning.zhang@uwindsor.ca).

Huici Wu is with the National Engineering Lab for Mobile Network Technologies, Beijing University of Posts and Telecommunications, Beijing 100876, China, and also with Peng Cheng Laboratory, Shenzhen 518066, China (e-mail: dailywu@bupt.edu.cn)

}

}


\maketitle

\begin{abstract}
Thanks to the advantages of flexible deployment and high mobility, unmanned aerial vehicles (UAVs) have been widely applied in the areas of disaster management, agricultural plant protection, environment monitoring and so on. With the development of UAV and sensor technologies, UAV assisted data collection for Internet of Things (IoT) has attracted increasing attentions. In this article, the scenarios and key technologies of UAV assisted data collection are comprehensively reviewed. First, we present the system model including the network model and mathematical model of UAV assisted data collection for IoT. Then, we review the key technologies including clustering of sensors, UAV data collection mode as well as joint path planning and resource allocation. Finally, the open problems are discussed from the perspectives of efficient multiple access as well as joint sensing and data collection. This article hopefully provides some guidelines and insights for researchers in the area of UAV assisted data collection for IoT.
\end{abstract}
\IEEEpeerreviewmaketitle

\begin{table*}[!t]
	\small
	\label{Table_1.1}
	\caption{Glossary.}
	\renewcommand{\arraystretch}{1.1} 
	\begin{center}

		\begin{tabular}{|m{0.8cm}m{4.3cm}|m{0.8cm}m{4.3cm}|m{0.8cm}m{4.3cm}|}
		\hline
		Abbr. & Definition & Abbr. & Definition & Abbr. & Definition\\
		\hline

		3D & Three-Dimensional.& DRL & Deep Reinforcement Learning. &MTCDs & Machine-Type Communication Devices.\\
    	\hline
   	
		5G & 5th Generation Mobile Networks.& FDMA & Frequency Division Multiple Access.&NOMA&Non-Orthogonal Multiple-Access.\\
		\hline
		
		ACO & Ant Colony Optimization. & G2A & Ground-to-Air.& PRM & Probabilistic Roadmap \\
		\hline
		
		AWGN & Additive White Gaussian Noise.& GA & Genetic Algorithm.& PSO & Particle Swarm Optimization.\\
		\hline
		
		AI & Artificial Intelligence.& GBD & Generalized Benders Decomposition.&QoS & Quality of Service.\\
		\hline

		AoI & Age of Information.&	GTOA & Group-based Trajectory Optimization Algorithm.&RL & Reinforcement Learning.\\
		\hline
		
		AP & Access Point.&HCP & Hybrid Clustering Routing Protocol.&RSSI &Rreceived Signal Strength Indicator.\\
		\hline
		
		BB & Branch and Bound.&Het-IoT & Heterogeneous Internet of Things.&SA & Simulated Annealing.\\
		\hline
		
		BCD & Block Coordinate Descent.&HHPS & Hybrid hovering points Selection.&SCA & Successive Convex Approximation.\\
		\hline
		
		BRB & Branch Reduction and Bound.&IoT &  Internet of Things.&SFLA & Shuffled Frog Leaping Algorithm.\\
		\hline

		BS & Base Station.&KKT & Karush-Kuhn-Tucker.&SIC & Successive Interference Cancelation.\\
		\hline
	
		CH & Cluster Head.&LEACH & Low Energy Adaptive Clustering Hierarchy.&STOA & Segment-based Trajectory Optimization Algorithm.\\
		\hline
		
		CP&Data collection Point.&LoS & Line of Sight.&TDMA & Time Division Multiple Access.\\
		\hline
		
		CRB & Cram\'{e}r-Rao Bound.&MA & Memetic Algorithm.&TSP& Traveling Salesman Problem.\\ 
		\hline
		
		CS & Cuckoo Search Algorithm.&MAC & Multiple Access Control.&TSPN & Traveling Salesman Problem with the Neighborhood.\\
		\hline
		
		DDPG & Depth Deterministic Gradient Descent.&MDP & Markov Decision Process. &UAV & Unmanned Aerial Vehicle. \\
		\hline
		
		DE & Differential Evolution.&MIMO & Multiple Input Multiple Output.&UKF& Unscented Kalman Filter. \\
		\hline
		
		DFP & Direct Future Prediction.&MLE & Maximum Likelihood Estimation.&VNS & Variable Neighborhood Search.\\
		\hline
		
		DL & Deep Learning.&MM & Majorize-Minimize or Minorize-Maximize.&		VoI & Value of Information. \\
		\hline
		
		DP & Dynamic Programming.&MPT & Microwave Power Transfer.& WSN & Wireless Sensor Network.\\
		\hline	
		\end{tabular}
	
	\end{center}
\end{table*}

\section{Introduction}

With the development of Internet of Things (IoT), sensors are widely deployed in the scenarios such as intelligent transportation, forest monitoring, smart city and smart ocean. It is estimated that by 2030, the number of sensors in the world will exceed 100 trillion \cite{Ehret2015The}. Hence, the collection of the huge amount of data from sensors faces great challenge. The data collection techniques of IoT require low energy consumption, low delay and high reliability.

\begin{figure}[!t]
	\centering
	\includegraphics[width=0.45\textwidth]{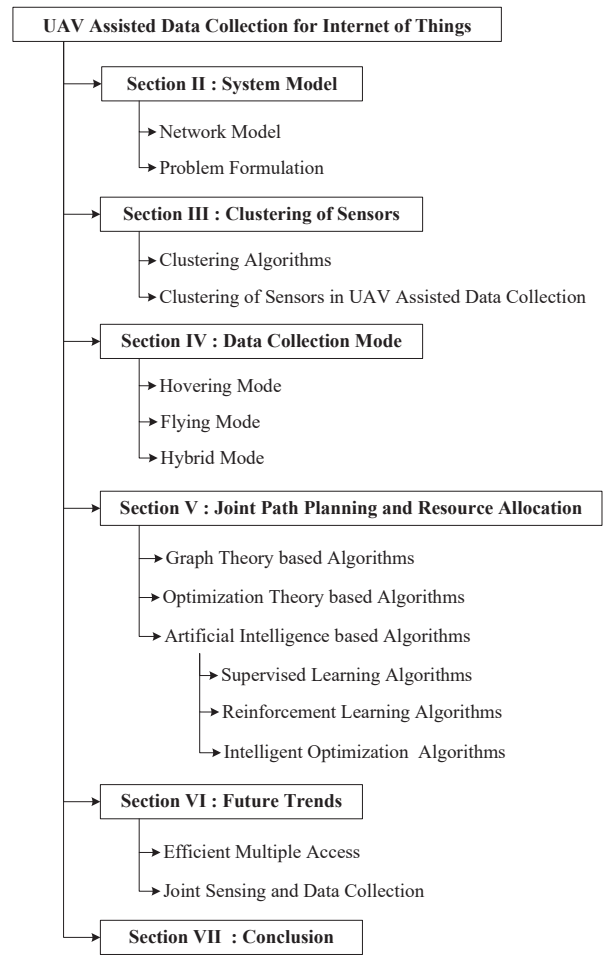}
	\caption{The organization of this article.}
	\label{fig_structure}
\end{figure}

Unmanned aerial vehicle (UAV) with wide coverage and high mobility provides new opportunities for data collection of IoT \cite{8417673,8468018}. UAV assisted data collection has the advantages of high agility, high flexibility and low cost. In addition, UAV can collect data in close proximity to sensors, which greatly reduces the energy consumption of IoT. The combination of UAV and IoT facilitates timely and effective data collection, especially in the complex, harsh or remote environments \cite{7036813,6564695,8207610,7854203,8647220}. As an ideal tool for data collection, UAV assisted data collection for IoT has the following advantages.

\begin{itemize}
	
	\item \emph{High efficiency and flexiblility of data collection:} UAV reduces the data collection time because its trajectory can be optimized \cite{8756665,8877250,8468195}.
	
	\item \emph{Low energy consumption and long network lifetime for IoT:} UAV not only reduces the energy consumption of data transmission in IoT, but also can charge the sensors wirelessly, which collectively improve the network lifetime of IoT \cite{8718663,8907457}.
	
	\item \emph{Wide coverage:} Due to the high altitude of UAV, there is a higher chance for line of sight (LoS) connections between UAV and sensors, which improves the success probability and coverage of communication \cite{8647220,7859281}.
	
\end{itemize}

Along with the above advantages, there are also several challenges in UAV assisted data collection. 

\begin{itemize}
	
	\item \emph{The clustering of sensors and selection of data collection mode:} The deployment and clustering of sensors, as well as the selection of data collection mode will greatly affect the performance of UAV assisted data collection. Hence, there exists a complex air-ground coupling in the optimization of the performance of data collection.
	
	\item \emph{Joint path planning and resource allocation of UAV:} The resource allocation problem is coupled with the path planning of UAV, which further complicates the problem of resource allocation.
	
\end{itemize}

Therefore, in view of the challenges of UAV assisted data collection, this article provides in-depth survey on three key technologies including the clustering of sensors, the modes of UAV assisted data collection, and the joint UAV path planning and resource allocation.

\begin{figure*}[!t]
	\centering
	\includegraphics[width=0.65\textwidth]{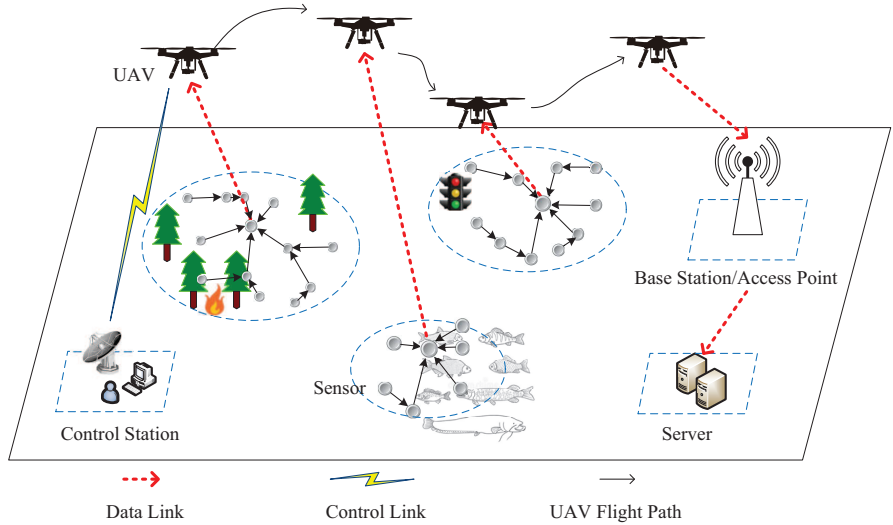}
	\caption{The system model of UAV assisted data collection.}
	\label{fig_scenarios}
\end{figure*}

\subsection{Existing Surveys and Tutorials}
Recently, a few related surveys and tutorials have been published. Li \emph{et al.} \cite{8579209} summarize the UAV communication for 5G/B5G wireless networks. They summarize and review the air-ground integrated network architecture and various 5G technologies implemented on UAV platform. In UAV assisted IoT, media access control (MAC) has a crucial impact on spectrum efficiency and energy efficiency of battery powered sensor networks. In \cite{8718663}, various MAC protocols for UAV assisted IoT are reviewed. Heaphy \emph{et al.} \cite{2017UAVs} describe the potential application of UAV in forestry. UAV makes up for the shortcomings of satellites and manned aircraft, and collects images with higher temporal and spatial resolution to support forest inventory, health monitoring, silviculture and harvesting operations. Pham \emph{et al.} \cite{2020Review} focus on the application of UAV in traffic data collection. This paper not only compares various UAV operation frameworks and popular platforms, but also applies UAV in speed behavior analysis, gap acceptance and merging behavior. In \cite{2019A}, Dan \emph{et al.} introduce the mode and latest progress of cooperation among various functional components in UAV assisted wireless sensor network (WSN). For large-scale IoT deployed in complex environment, Yang \emph{et al.}  \cite{9251117} review four technologies, including sensor deployment, UAV configuration (such as flight speed and radius), UAV path planning and UAV autonomous navigation. However, existing articles have the following limitations.

\begin{itemize}
	
	\item There is a lack of investigation and discussion on the data collection modes using UAV.
	
	\item The clustering algorithms of sensors are not discussed in details.
	
	\item The joint path planning and resource allocation schemes for UAV assisted data collection are not reviewed.
	
\end{itemize}

\subsection{Contributions and Organization}

This article aims to review the development status and future trends of UAV assisted data collection techniques. To this end, we conduct a comprehensive review on the relevant literatures in recent years. The scenario of UAV assisted data collection is not a simple combination of UAV networking and IoT. They need to cooperate with each other to improve the system efficiency, which is the perspective of this article. The main contributions of this article are summarized as follows.

\begin{itemize}
	\item Three key technologies of UAV assisted data collection for IoT including the clustering of sensors, the data collection modes, and the joint UAV path planning and resource allocation methods, are reviewed in details.
	\item The complete solution of UAV assisted data collection including the key technologies such as joint sensing and data collection, clustering of sensors, efficient access of large-scale sensors, and efficient routing in WSN, and joint path planning and resource allocation, is designed. Among these key technologies, efficient multiple access and joint sensing and data collection have a large probability to be the future research trends. 
\end{itemize}

As shown in Fig. \ref{fig_structure}, the rest of this article is organized as follows. Section II introduces the system model of UAV assisted data collection for IoT. In Section III, the clustering algorithms of sensors are introduced. Section IV investigates three data collection modes, analyzing the advantages and disadvantages of each data collection mode. Section V reviews the joint UAV path planning and resource allocation methods in details. Section VI discusses the complete solution and future trends of UAV assisted data collection. Finally, Section VI summarizes this article.

\section{System Model}

In this section, we introduce the network model and mathematical model in UAV assisted data collection.

\subsection{Network Model}

As illustrated in Fig. \ref{fig_scenarios}, the network elements of UAV assisted data collection consist of sensor, UAV, control station and server, which are explained as follows.

\begin{itemize}
	
	\item \emph{Sensor:} In order to sense the environment, the sensors are deployed and the sensing data are sent to UAV. The type of sensors depends on the application scenarios. For instance, the temperature and humidity sensors are deployed in the forest monitoring scenario.
	\item \emph{UAV:} The UAV flies above the sensors to collect data, which is then brought back to the server.
	\item \emph{Control station:} The control station plans UAVs' flight path for efficient data collection.
	\item \emph{Base station (BS)/Access point (AP):} The BS/AP collects the sensing data directly from the surrounding sensors, as well as the UAVs.
	\item \emph{Server:} The server stores and processes the data from sensors.

\end{itemize}

As shown in Fig. \ref{fig_scenarios}, in the scenario of UAV assisted data collection, UAV takes off from a control station and flies above sensors to collect data, such as the potential of hydrogen (PH) and salinity in the ocean, the wildlife and fire in the forest, the traffic and weather in the smart city. Then the UAV delivers the data to the BS/AP and finally returns to the control station. The scheme of UAV assisted data collection aims at high efficiency, low energy consumption and low delay during data collection.

In addition, some standard IoT technologies are applied in UAV assisted data collection, such as NB-IoT and LoRa. LoRa can be used as a gateway for UAV assisted data collection to extend the coverage to rural and remote areas \cite{8965135}. Delafontaine \emph{et al.} \cite{9196869} present a UAV assisted localization system using LoRa networks in which a UAV is deployed to improve the positioning accuracy of sensor. Because NB-IoT can support large coverage  and meet the low-power and low-cost operation of sensors, it is also applied in UAV assisted data collection, such as soil detection \cite{9042335}, city open water monitoring \cite{9265880} and air quality monitoring \cite{8675167}.

\begin{figure}[!t]
	\centering
	\includegraphics[width=0.5\textwidth]{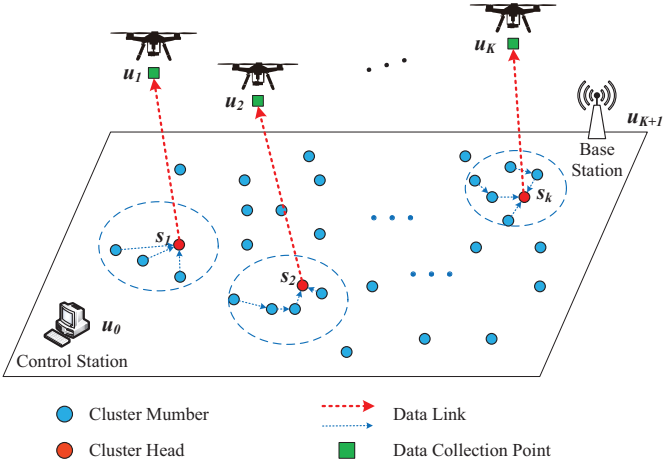}
	\caption{Optimization of UAV assisted data collection.}
	\label{fig_mathematical_model}
\end{figure}

\subsection{Mathematical Model}

As shown in Fig. \ref{fig_mathematical_model}, the system of UAV assisted data collection is composed of a UAV, a BS, a control station and $M$ sensors. The set of sensors is denoted by ${\cal M}  \buildrel \Delta \over =  \left\{ {1,2, \ldots ,M} \right\}$. The UAV collects and sends the data of sensors back to the BS. The control station plans the flight path of the UAV and charges the UAV before flight. There are the following assumptions \cite{8540462,8613833}.

\begin{itemize}
	
	\item \emph{Distribution of sensors:} Sensors are randomly distributed in the target area. Their position coordinates are known and expressed as ${s_m} = {\left( {{x_m},{y_m},h} \right)^T} \in {\mathbb{R}^{3 \times 1}},m\in{\cal M}$. The sensor is equipped with omni-directional antenna and the communication range is ${r_c}$.
	\item \emph{Clustering of sensors:} When the number of sensors is small or the locations are scattered, each sensor can directly upload the data to the UAV. When the number of sensors is large, the sensors need to be clustered. Hence there are two types of nodes, namely, cluster head (CH) and cluster member (CM). The CH collects the data of CMs within the cluster, and is responsible for communicating with the UAV \cite{9014306}.
	\item \emph{Self-positioning of UAV:} UAV acquires its location via the installed positioning module such as Global Navigation Satellite System (GNSS) module \cite{8502536}.
	\item \emph{Number and type of UAVs:} The number of UAVs depends on the number of sensors and the amount of data to be collected. The type of UAVs consists of rotor UAV and fixed-wing UAV \cite{2006Flight}. Therefore, three data collection modes, i.e., hovering mode, flying mode and hybrid mode, exist in the literatures. 
	\item \emph{Mobility, energy and capacity of UAVs:} UAV can fly at a fixed altitude $H$ and speed $v$ (When $v = 0$, it means that the UAV is hovering), and $V_{max}$ is the maximum speed that UAV can reach. The maximum flight distance and energy of UAV are $L_{max}$ and $E_{max}$, respectively \cite{8489991}. The buffer size of UAV is also taken into account, limiting the amount of data that can be collected.
	\item \emph{Ground-to-air communication and transmission rate:} There will be LoS channel between sensors and UAV with a certain probability. This probability depends on the environment and the angle between the sensor and the UAV. As shown in \cite{8613833,9145752}, the probability of LoS transmission is 
	\begin{equation}\label{2.1}
		\begin{aligned}
			P_{\rm{LoS}}\left( {{s_m},{u}} \right) = \frac{1}{{1 + a\exp \left[ { - b\left( {\theta \left( {{s_m},{u}} \right) - a} \right)} \right]}},
		\end{aligned}
	\end{equation}
	where ${\theta \left( {{s_m},{u}} \right)}$ is the angle between the sensor ${s_m},m \in {\cal M}$ and the UAV at position $u \in {\mathbb{R}^{3 \times 1}}$. $a$ and $b$ are environmental parameters, which mainly depend on the ratio of building area to total land area, the average number of buildings in a unit area and the height of buildings \cite{8613833}. Then the transmission rate between the sensor ${s_m}$ and the UAV located at $u$ is \cite{8779596}
	\begin{equation}\label{2.2}
	\begin{aligned}
		R\left( {{s_m},u} \right) = B{\log _2}\left( {1 + \frac{{{{\left| {h\left( {{s_m},u} \right)} \right|}^2}{P_m}}}{{{N_0}}}} \right),
	\end{aligned}
	\end{equation}		
	where $B$ is the channel bandwidth, ${h\left( {{s_m},u} \right)}$ is the channel gain between UAV and the sensor, $P_m$ is the transmit power of each sensor when communicating with UAV, $N_0$ is the power of additive Gaussian white noise.
\end{itemize}

Combined with the model in Fig. \ref{fig_mathematical_model} and the above assumptions, it is assumed that $M$ sensors in the IoT are divided into $K$ clusters, expressed as ${\cal K} \buildrel \Delta \over =  \left\{ {1,2, \ldots ,K} \right\}$. Each cluster has only one CH $s_k$, and the sensors in the cluster can communicate with each other. The goal of UAV is to collect data from the CHs quickly and efficiently. In order to achieve this goal, the distance, flight time, and energy consumption for path planning need to be jointly optimized \cite{9251117}. Meanwhile, the data collection mode between UAV and CH, such as hovering mode, flying mode and hybrid mode, has an impact on the efficiency of data collection \cite{8756751}. Taking hovering mode as an example, the UAV collects data of CH $s_k$ at the data collection point $u_k,k \in {\cal K}$ \cite{8432487}. Other data collection modes have corresponding constraints. Before formulating the problem, the following constraints need to be considered.

\subsubsection{Sensor energy  consumption constraint}
This constraint ensures that the energy consumption of each sensor $s_k$ does not exceed $E_k$.
	\begin{align}
			{t_k}{P_k} &\le {E_k}{\kern 1pt} {\kern 1pt} {\kern 1pt} {\kern 1pt} {\kern 1pt} {\kern 1pt} {\kern 1pt} {\kern 1pt} {\kern 1pt} {\kern 1pt} \forall k \label{2.5.1},\\
			{P_k} & >  0{\kern 1pt} {\kern 1pt} {\kern 1pt} {\kern 1pt} {\kern 1pt} {\kern 1pt} {\kern 1pt} {\kern 1pt} {\kern 1pt} {\kern 1pt} {\kern 1pt} \forall k \label{2.5.2}.
	\end{align}
In hovering mode, the channel conditions between UAV and sensors remain unchanged, hence the sensor $s_k$ can transmit data with constant transmit power $P_k$. $t_k$ is the hovering time of UAV at data collection point $u_k$.
\subsubsection{Data collection constraint}
The sensor $s_k$ needs to meet the minimum data collection requirements $r_k$.

\begin{equation}\label{2.6}
	\begin{aligned}
		{t_k}B{\log _2}\left( {1 + \frac{{{{\left| {h\left( {{s_k},{u_k}} \right)} \right|}^2}{P_k}}}{{{N_0}}}} \right) \ge {r_k}{\kern 1pt} {\kern 1pt} {\kern 1pt} {\kern 1pt} {\kern 1pt} {\kern 1pt} {\kern 1pt} {\kern 1pt} {\kern 1pt} {\kern 1pt} \forall k.
	\end{aligned}
\end{equation}
\subsubsection{UAV energy  consumption constraint}
The total energy consumption of UAV cannot exceed its maximum energy $E_{\max}$.
\begin{equation}\label{2.7}
	\begin{aligned}
		{P_u}\left( 0 \right)\sum\limits_{k = 1}^K {{t_k}}  + {P_u}\left( v \right)\frac{{\sum\limits_{i = 0}^{K + 1} {\sum\limits_{j = 0}^{K + 1} {{x_{i,j}}d{}_{i,j}} } }}{v} \le {E_{\max}},
	\end{aligned}
\end{equation}
where ${P_u}\left( v \right)$ is the power of UAV flying at constant speed $v$ ($0 < v \le {V_{\max }}$) \cite{9447204,8119562}, ${d_{i,j}}$ is the distance from $u_i$ to $u_j$.
\subsubsection{UAV trajectory constraint}
This constraint ensures that the UAV starts from the control station, traverses all data collection points and then returns to the BS.
\begin{align}
	\sum\limits_{i = 1}^K &{{x_{i,j}}} = 1{\kern 1pt} {\kern 1pt} {\kern 1pt} {\kern 1pt} {\kern 1pt} {\kern 1pt} {\kern 1pt} {\kern 1pt} {\kern 1pt} {\kern 1pt}  i \ne j,\forall j \in {\cal K}  \label{2.8.1},\\
	\sum\limits_{j = 1}^K &{{x_{i,j}}} = 1{\kern 1pt} {\kern 1pt} {\kern 1pt} {\kern 1pt} {\kern 1pt} {\kern 1pt} {\kern 1pt} {\kern 1pt} {\kern 1pt} {\kern 1pt}  i \ne j,\forall i \in {\cal K} \label{2.8.2},\\
	\sum\limits_{j = 1}^K &{{x_{0,j}}} = \sum\limits_{i = 1}^K {{x_{i,K + 1}}} = 1{\kern 1pt} {\kern 1pt}  {\kern 1pt} {\kern 1pt} {\kern 1pt} {\kern 1pt} {\kern 1pt}  \forall i,j \in {\cal K}\label{2.8.3},\\
	{{x_{i,j}}},{x_{0,j}},&{x_{i,K + 1}} \in \left\{ {0,1} \right\}{\kern 1pt} {\kern 1pt}i \ne j,\forall i,j \in {\cal K} \label{2.8.4},
\end{align}
where ${x_{i,j}}$ is a binary variable. ${x_{i,j}}=1$ indicats that the UAV flies from the data collection point $u_i$ to $u_j$. $u_0$ and $u_{K+1}$ represent the starting point (control station) and ending point (BS) of the UAV, respectively.
\subsubsection{Age of information constraint}
The age of information (AoI) is a performance indicator to measure the timeliness of data. Compared with the latency, AoI includes not only the transmission delay of data packets, but also the waiting time of data packets at the source node and the residence time at the destination node \cite{9380899,9312959}. AoI constraint reveals the requirements for data timeliness.
\begin{equation}\label{2.9}
	\begin{aligned}
		A_j^k \le A_j^{k,th}{\kern 1pt} {\kern 1pt} {\kern 1pt} {\kern 1pt} {\kern 1pt} {\kern 1pt} {\kern 1pt} \forall k \in {\cal K},j \in {{\cal J}_{k}},
	\end{aligned}
\end{equation}
where ${A_j^k}$ and $A_j^{k,th}$ are the AoI and its threshold for the $j$-th uploaded packet of CH $s_k$, respectively. ${{\cal J}_k} \buildrel \Delta \over = \left\{ {1,2, \ldots ,{J_k}} \right\}$ is the number of data packets in $s_k$. Without considering the sampling time and communication overhead, the AoI of the data collected at CH $s_k$ consists of the time of data upload, the time of UAV returning to the data center and the time of data unloading \cite{8756665,9681851}. If the UAV accesses CHs sequentially (from $s_1$ to $s_K$), the AoI of the $j$-th uploaded packet from the CH $s_k$ can be expressed as \cite{8756665}
\begin{equation}\label{2.10}
	\begin{aligned}
		A_j^k = \sum\limits_{i = k}^K {\sum\limits_{j = 1}^{{J_k}} {\varsigma _j^i} }  - \sum\limits_{l = 1}^{j - 1} {\varsigma _l^k}  + {\varsigma _0} + \frac{{\sum\limits_{i = k}^K {\left\| {{u_i} - {u_{i + 1}}} \right\|} }}{v},
	\end{aligned}
\end{equation}
where $\varsigma _j^i$ is the time of uploading the $j$-th data packet for the $i$-th CH, ${\varsigma _0}$ is the data unloading time of UAV. For the data packets of CHs not accessed by UAV, the AoI is $0$. If further considering the possibility of packet expiration, AoI can be expressed as \cite{8752017}
\begin{equation}\label{2.11}
	\begin{aligned}
		A_j^k = \left\{ \begin{array}{l}
			\sum\limits_{i = k}^K {\sum\limits_{j = 1}^{{J_k}} {\varsigma _j^i} }  - \sum\limits_{l = 1}^{j - 1} {\varsigma _l^k}  + {\varsigma _0} + \frac{{\sum\limits_{i = k}^K {\left\| {{u_i} - {u_{i + 1}}} \right\|} }}{v},t_j^k < t_j^{{{k,ex}}}\\
			{T_{{\rm{total}}}},{\rm{otherwise}},
		\end{array} \right.
	\end{aligned}
\end{equation}
where $t_j^k$ is the time interval from the generation of the $j$-th data packet of $s_k$ to the completion of data collection by UAV, $t_j^{{{k,ex}}}$ is the effective time of the data packet. If the data packet expires, its AoI is set as the overall time for the UAV to complete the task. 

The above parameters about queuing delay need to be modified according to the queuing models, such as First-Come-First-Service (FCFS), Last-Come-First-Service (LCFS), priority based and limited buffer strategies \cite{6195689,7415972}. The application scenarios of FCFS, LCFS and priority based model are increasingly sensitive to data timeliness. The scenarios requiring comprehensive data such as agricultural monitoring and marine monitoring often adopt the FCFS model. The LCFS model is more suitable for the scenarios that value the latest data, such as driverless vehicle monitoring and field fire monitoring.

UAV assisted data collection schemes usually aim at maximizing the amount of collected data, minimizing flight time of UAV, minimizing energy consumption of UAV or sensor, minimizing AoI and so on. Overall, there are three key issues in UAV assisted data collection, namely, the clustering of sensors, the data collection mode of UAVs, and the path planning and resource allocation of UAVs. The application and optimization of the three key technologies are helpful to improve the efficiency of data collection, ensure the timeliness of data, as well as save the energy consumption of UAV and sensors. Besides, it is essential to consider the clustering algorithm of sensors and the data collection mode of UAV for an effective path planning algorithm. In the following sections, we will provide a thorough survey on these topics. 

\section{Clustering of Sensors }

In this section, we first review the clustering algorithms in conventional sensor network, and then discuss the clustering of sensors in UAV assisted data collection.

\subsection{Clustering Algorithms}

\begin{table*}[!t]
	\label{Table_3.1}
	\small
	\caption{Factors affecting clustering algorithms.}
	\renewcommand{\arraystretch}{1.3} 
	\begin{center}
		
		\begin{tabular}{|m{0.08\textwidth}<{\centering}|m{0.35\textwidth}<{\centering}|m{0.08\textwidth}<{\centering}|m{0.35\textwidth}<{\centering}|}
			\hline
			Reference & Methodologies & Reference & Methodologies\\
			
			\hline
			\cite{8927929} &  Select the sensor with the highest power as the cluster head (CH). & \cite{8540462} & Distributed clustering.\\
			\hline
			
			\cite{8756665} & Affifinity propagation. & \cite{8647924,9014306} & \textit{K}-means clustering algorithm. \\
			\hline

			\cite{6549957,8613833} & Problem decomposition. & \cite{2019Unmanned} & $\alpha$-hop clustering algorithm.\\
			\hline
			
			\cite{8881889} & Base station (BS) assisted clustering. & & \\
			\hline
			
		\end{tabular}
	\end{center}
\end{table*}

When the number of sensors is large, the topology of sensor network becomes complicated and the cost of routing between sensors is relatively large. Therefore, it is necessary to utilize a clustering algorithm to divide the network into several sub-networks. The basic idea of clustering algorithms is to merge neighboring nodes into a cluster and select a CH from each cluster that is responsible for aggregating the data from other sensors in the cluster.

Depending on different performance metrics, there are various clustering algorithms. According to the selection methods of CH, the clustering algorithms are classified into deterministic CH selection algorithms, random CH selection algorithms and adaptive CH selection algorithms \cite{7976279}. According to the control methods of clustering algorithms, the clustering algorithms can be classified into distributed clustering algorithms, hybrid clustering algorithms and centralized clustering algorithms \cite{2017Survey}.

The widely applied clustering protocol is the low energy adaptive clustering hierarchy (LEACH) because of its simplicity \cite{7855660}. More specifically, at the beginning of LEACH, each sensor generates a random number between 0 and 1. If the number is greater than a given threshold, it is selected as CH. Then, the CH broadcasts a packet to other sensors in the same network. Each node decides which cluster to join according to the strength of the received signal, and transmits a feedback packet to the CH in a time division multiple access (TDMA) manner. With LEACH, the energy consumption of each sensor is distributed evenly by periodically switching the CHs, thereby prolonging network lifetime \cite{7976279}. Recently, there are some literatures to improve LEACH \cite{0Leach,7855660,2015LEACH}. Basavaraj and Jaidhar \cite{8358338} propose a new threshold by calculating the average remaining energy of sensors, the average distance between sensors and BS, and the optimal number of CHs. In the scenario of UAV assisted data collection, since LEACH randomly selects CH, some sensors with less energy may be selected, so that the communication with the UAV cannot be completed. Chen and Shen \cite{9532391} comprehensively consider the factors such as the residual energy and the communication range of sensor to select CH, which extend the network life. The malicious node may be selected as CH under LEACH. Therefore, Wang \emph{et al.} \cite{8458023} improve LEACH and apply UAV to collect sensor information such as energy and ID, so as to avoid selecting the sensor affected by malicious node as CH.

\subsection{Clustering of Sensors in UAV Assisted Data Collection}

In the scenario of UAV assisted data collection, due to the limited energy of UAV, the energy consumption brought by the flight of UAV needs to be reduced. A large number of clusters will reduce the energy consumption within the cluster to forward data, but will increase the energy consumption of UAV for flight. In contrast, a small number of clusters will reduce the energy consumption of UAV during flight, but the energy consumption of sensors will increase. Therefore, it is necessary to jointly optimize the energy consumption of sensors, the energy consumption of UAV, clustering of sensors and CH selection, path planning of UAV \cite{8647924}. The clustering methods of sensors used in relevant literatures are shown in Table \uppercase\expandafter{\romannumeral2}.
\begin{table*}[!t]
	\label{Table_4.1}
	\small
	\caption{Comparison of data collection modes.}
	\begin{center}
		\begin{threeparttable}
			\renewcommand{\arraystretch}{1.1} 

			\begin{tabular}{|m{0.15\textwidth}<{\centering}|m{0.15\textwidth}<{\centering}|m{0.15\textwidth}<{\centering}|m{0.12\textwidth}<{\centering}|m{0.12\textwidth}<{\centering}|m{0.15\textwidth}<{\centering}|}
				\hline
				Data collection mode & Hovering at data collection point (CP) &  Continuing data collection after passing through CP & Energy consumption & Flight time & The AoI of data collected from sensors\\
				
				\hline
				Hovering mode & $\surd$ & $\times$ & High & High & Low\\
				\hline
				Flying mode & $\times$ & $\surd$ & Low & Low & High\\
				\hline
				Hybrid mode & $\surd$ & $\surd$ & Medium & Medium & Medium\\
				\hline
				
			\end{tabular}
			
			\begin{tablenotes}
				\footnotesize
				\item[1] $\surd$ denotes the existence of the feature, $\times$ denotes the absence of the feature 
			\end{tablenotes}
			
		\end{threeparttable}
	\end{center}
\end{table*}

\begin{table*}[!t]
	\label{Table_4.2}
	\small
	\caption{The optimization schemes under different data collection modes.}
	\renewcommand{\arraystretch}{1.1} 
	\begin{center}
		
		\begin{tabular}{|m{0.2\textwidth}<{\centering}|m{0.1\textwidth}<{\centering}|m{0.25\textwidth}<{\centering}|m{0.35\textwidth}<{\centering}|}
			\hline
			Data collection mode & Reference & Objective function & \makecell[c]{Decision variables}\\
			\hline
			
			\multirow{4}[12]{\linewidth}{\makecell[c]{Hovering mode}}
			& \cite{8743453} & Minimization of the total task duration. &The number of subregions, hovering point,   hovering time of each position, trajectory between hovering points.\\
			\cline{2-4}
			& \cite{9014306} & Minimization of flight time of UAV.&The number and trajectory of UAV.\\
			\cline{2-4}
			& \cite{8907457} & Maximum of minimum energy of sensors after data collection.&Hovering point, residence time of UAV.\\
			\cline{2-4}
			& \cite{8756665} & Minimization of node upload time and flight time of UAV.&Position correlation between sensors and hover points, trajectory of UAV.\\
			\hline
			
			\multirow{2}[10]{\linewidth}{\makecell[c]{Flying mode}}
			& \cite{8377357} &Minimization of flight time of UAV.&Continuous optimization of interval, speed of UAV, transmit power of sensors.\\
			\cline{2-4}
			& \cite{9013148} &Minimization of the weighted sum of energy consumption of UAV and sensors.&Trajectory of UAV, wake-up scheduling of sensors, data collection time.\\
			\cline{2-4}
			\hline
			
			\multirow{2}[6]{\linewidth}{\makecell[c]{Hovering and \\ Flying mode}}
			& \cite{8779596} &Minimum of maximum data collection time.&Trajectory of UAV, wake-up scheduling of sensors, sensor association.\\
			\cline{2-4}
			& \cite{8432487} &Minimization of flight time of UAV.&Data collection interval, speed of UAV, transmit power of sensors.\\
			\cline{2-4}
			\hline
			
			\multirow{2}[6]{\linewidth}{\makecell[c]{Hybrid mode}}
			& \cite{8644379} &Minimization of battery consumption and Maximization of UAV throughput.&Energy consumption and throughput of UAV, delay of machine-type
			communications devices (MTCDs) tasks, collection and computing efficiency of different priorities.\\
			\cline{2-4}
			& \cite{8756751} &Minimization of the average age of information (AoI).&The age of collected data, data collection mode, energy consumption of each sensor.\\
			
			\hline
		\end{tabular}
	\end{center}
\end{table*}

Chen \emph{et al.} \cite{8927929} and Salih \emph{et al.} \cite{8540462}  consider the energy constraints of sensors for clustering. Chen \emph{et al.} \cite{8927929} select the sensor with the highest power as the CH to extend the lifetime of sensors. The data forwarding rule in the cluster is that only when the value of information (VoI), which is highest when an event first occurs and decays with time, is greater than a certain threshold, the sensors in the cluster will send the data to the CH, thereby greatly reducing the energy consumption of the sensors. Salih \emph{et al.} \cite{8540462} utilize a distributed clustering method to minimize the average sending distance from the sensors in the cluster to the CH. In order to minimize the AoI, Tong and Moh \cite{8756665} jointly optimize the CH selection and flight path of UAV, in which an algorithm based on affinity propagation is used to determine the locations of the CH and the associated sensor. 

In addition, most of the literatures considers the energy consumption of sensors, flight time of UAV and path planning in the clustering algorithms. Ebrahimi \emph{et al.} \cite{8647924} and Alfattani \emph{et al.} \cite{9014306} apply the \textit{k}-means clustering algorithm, and update the sensors in the cluster through iterations. Sujit \emph{et al.} \cite{6549957} and Ghorbel \emph{et al.} \cite{8613833}  decompose the joint UAV path planning and energy consumption optimization problem into several sub-problems. In \cite{6549957}, the initial problem is decomposed into the sub-problems of finding clusters, connecting clusters, route planning, and UAV path planning. Ghorbel \emph{et al.} \cite{8613833} decompose the problem into three sub-problems: optimizing the position of CH, assigning sensors into clusters and UAV path planning. The $\alpha$-hop clustering algorithm designed by Wu \emph{et al.} \cite{2019Unmanned} adjust the hop number in each cluster to reduce energy consumption and delay. A large $\alpha$ determines small number of clusters, which reduces the UAV path and decreases the flight time. When $\alpha$ is relatively small, each cluster is small, which reduces energy consumption of data forwarding in sensor network. Haider \emph{et al.} \cite{8881889} apply UAV to get the remaining energy of sensors, channel conditions and the distance between sensors. Then the BS utilizes these information from UAV and determines which sensors are CHs, which greatly extends the lifetime of the sensors.

\section{Data Collection Mode}

According to the movement status of UAV during data collection, the data collection modes are classified into three types, namely, hovering mode, flying mode and hybrid mode. Their characteristics are summarized in Table \uppercase\expandafter{\romannumeral3}. Table \uppercase\expandafter{\romannumeral4} is a summary of the literatures on the three data collection modes.
\subsection{Hovering Mode}

In hovering mode, the UAV stays above the data collection point for a period of time to collect data. The data collection point can be the location of the CH or near the CH. In \cite{8743453}, the entire region is divided into several sub-regions, and UAV hovering over these sub-regions collects data from sensors. Hovering mode can provide relatively stable data transmission, but the energy consumption is relatively large.

The UAV hovering locations and duration have a great impact on the reliable data collection and the shortest flight time. A framework design for IoT based on multiple UAVs is proposed in \cite{8756751}. This work firstly optimize the number and location of CHs to minimize the data collection time. Then, the number of UAVs and their trajectories are optimized in order to minimize the flight time. And reliable data collection shortens the AoI of data collected from sensors, which is the time that the data is transmitted from sensor to the data center. In \cite{8756665}, in order to minimize the maximum AoI of all sensors, a joint sensor association and path planning strategy is proposed to balance the flight time of UAV and the upload time of sensors which is affected by the quality of ground-to-air communication link under hovering mode.

In addition, to maximize the lifetime of sensors, the UAV hovering height and duration should be considered. Under the constraints of transmit power of sensors and energy consumption of UAV, Baek \emph{et al.} \cite{8907457} optimize the hovering point and dwell time of UAV to maximize the minimum energy of sensors during data transmission and energy collection periods.

\begin{table*}[!t]
	\label{Table_5_1}
	\small
	\caption{Comparison of graph theory based algorithms}
	\begin{threeparttable}
		\renewcommand{\arraystretch}{1.1} 
		\begin{center}
			
			\begin{tabular}{|m{0.1\textwidth}<{\centering}|m{0.08\textwidth}<{\centering}|m{0.2\textwidth}<{\centering}|m{0.22\textwidth}<{\centering}|m{0.05\textwidth}<{\centering}|m{0.05\textwidth}<{\centering}|m{0.05\textwidth}<{\centering}|m{0.05\textwidth}<{\centering}|}
				\hline
				Algorithm & Reference & Objective function & Decision variables & AoCD & TE & EE & AoI\\
				\hline
				
				\multirow{2}[6]{\linewidth}{\makecell[c]{Voronoi \\ diagram}}
				& \cite{8933037} & Maximum of minimum residual energy of sensors. & Hovering point, UAV path planning. & $\times$ & $\times$ & $\surd$ & $\times$ \\
				\cline{2-8}
				& \cite{9120646} & Minimization of flight time of UAV.& UAV task area allocation, UAV path planning. & $\times$ & $\surd$ & $\surd$ & $\times$\\
				\cline{2-8}
				\hline
				
				\multirow{1}[1]{\linewidth}{\makecell[c]{Probabilistic \\ roadmap}}
				& \cite{8741059} & Maximization of the amount of collected data. & UAV path planning. & $\surd$ & $\times$ & $\times$ & $\times$ \\
				\cline{2-8}
				\hline
				
				\multirow{1}[1]{\linewidth}{\makecell[c]{Hilbert \\ curve}}
				& \cite{8943139} & Maximization of data collection rate. & UAV path planning. & $\surd$ & $\times$ & $\times$ & $\times$ \\
				\cline{2-8}
				
				\hline
			\end{tabular}
			\begin{tablenotes}
				\footnotesize
				\item[1] $\surd$ denotes the existence of the feature, $\times$ denotes the absence of the feature
				\item[2] AoCD: Amount of collected data, TE: Time efficiency, EE: Energy efficiency, AoI: Age of information
			\end{tablenotes}
		\end{center}
		
	\end{threeparttable}
\end{table*}
\subsection{Flying Mode}

Under flying mode, when UAV passes through the data collection point, it slows down and collects the data of sensors until it leaves the communication range of the sensors. With the characteristics of moving while collecting data in flying mode, the collection task can be completed quickly. However, if the amount of data is large, the UAV in flighting mode may not satisfy the the requirement of timely data collection.

Under flying mode, the shortest flight time is commonly chosen as the objective function. In UAV assisted data collection, it is observed that the optimal speed of UAV is proportional to the energy and density of sensors. However, it is inversely proportional to the amount of data to be uploaded. Thus, to minimize the flight time of UAV during data collection, the horizontal distance that the sensors can upload data to the UAV, the speed of UAV, and the transmit power of sensors are jointly optimized in \cite{8377357} and \cite{8432487}. In the multi-UAV enabled IoT, Zhan \emph{et al.} \cite{8779596} focus on the problem of minimizing the maximum completion time of data collection among all UAVs. They jointly optimize the trajectories of UAVs, wake-up scheduling and association for sensors, while ensuring that sensors can successfully upload a given amount of data with limited energy budget.

In the above literatures, the constraint is that the energy of UAV or sensor is limited. Zhan and Huang \cite{9013148} study the fundamental tradeoff between the energy consumption of UAV and that of all sensors. In order to characterize this tradeoff relation, they formulate an optimization problem to minimize the weighted sum of the energy consumption of UAV and sensors. Then, the trajectory of UAV, mission completion time and wake-up scheduling for all sensors are jointly optimized.

\subsection{Hybrid Mode}
The hybrid mode combines the characteristics of hovering mode and flying mode. The UAV stays above the data collection point for data collection. When the UAV flies to the next collection point after a period of time, data collection can still be carried out until the UAV is out of the communication range of the sensor. The energy consumption and flight time under hybrid mode are smaller than that under hovering mode, but larger than that under flying mode. Different data collection modes are usually chosen according to the application scenarios and performance requirements.

The hybrid mode can adjust the data collection mode dynamically according to the energy consumption of UAV. To minimize the average AoI of data collection, Jia \emph{et al.} \cite{8756751} lower the AoI of data collection under three data collection modes, energy consumption at each node and age evolution of collected data. When the initialized energy of sensor increases, hybrid mode has better performance than the other modes. Because of the limited battery of machine-type
communications devices (MTCDs) and UAV, Zhu \emph{et al.} \cite{8644379} propose a hybrid hovering points selection (HHPS) algorithm to select the hovering points of UAV with the minimum power consumption of MTCDs.

\section{Joint path planning and resource allocation}

Path planning of UAV refers to the calculation of the optimal path of UAV from the source to the destination under specified conditions \cite{2010Research}. The UAV path planning has a crucial influence on the performance of data collection. There are various goals for path planning, such as the shortest path, the shortest flight time and the lowest energy consumption of UAV. The characteristics of UAV, such as high mobility and flexible deployment, bring challenges for UAV path planning. Besides, considering the limited energy and radio resources of UAV, the UAV path planning is usually combined with resource allocation to efficiently optimize the resource utilization \cite{9057370,8345703}. Hence, the complexity of joint path planning and resource allocation is high because the decision space is huge.

In this section, according to the techniques in the algorithms, the updating methods, objective functions and so on, we classify the path planning algorithms into graph theory based algorithms, optimization theory based algorithms and artificial intelligence (AI) based algorithms. Then, the relevant literatures of UAV path planning are reviewed in the scenario of UAV assisted data collection.

\subsection{Graph Theory based Algorithms }

Graph theory based path planning algorithms include Voronoi diagram algorithm, probabilistic roadmap (PRM) algorithm, Hilbert curve algorithm, and so on. Using graph theory based algorithm, the geographical space where UAV is flying is converted into a graph and the path between source and destination is searched. The comparison of different algorithms is listed in Table \uppercase\expandafter{\romannumeral5}, where the key metrics are explained as follows.

\begin{itemize}
	
	\item \emph{Amount of collected data (AoCD):} The amount of data collected by UAV.
	\item \emph{Time efficiency (TE):} The time duration for UAV to complete data collection.
	\item \emph{Energy efficiency (EE):} The energy consumption of UAV and sensors in terms of battery or fuel. 
	\item \emph{Age of information (AoI):} Freshness of collected data, which needs to be distinguished from the latency and is defined as the time interval between the time when the sensor data is generated and the time when the UAV returns and uploads the collected data to the data center. Specifically, the latency is the time interval between the sending time and the receiving time of data, while AoI is the time interval between the generation time and the use time of data \cite{9380899,9312959,9681851}.
	
\end{itemize}

In UAV assisted data collection, AoCD, TE, EE, AoI are the key performance metrics that may be considered in the algorithm design.

\begin{figure}[!t]
	\centering
	\includegraphics[width=0.3\textwidth]{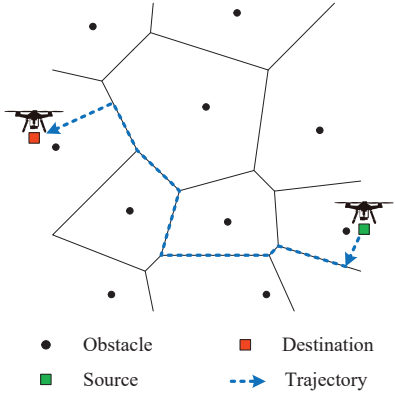}
	\caption{The principle diagram of Voronoi diagram.}
	\label{fig_voronoi}
\end{figure}

\subsubsection{Voronoi diagram}

Voronoi diagram, also known as Tyson polygon or Dirichlet diagram, is a set of polygons composed of vertical bisectors connecting two adjacent points, where the points are the obstacles \cite{6852323,2006VORONOI}. The advantage of Voronoi diagram is that the generated path is far away from obstacles. Thus collision can be avoided. As shown in Fig. \ref{fig_voronoi}, the circular is regarded as the obstacle and the plane is divided in light of the nearest neighbor principle. The point within a polygon is nearest to the obstacle within this polygon compared with other obstacles. Then, the shortest path is searched along the edges of the polygons as the flight path of UAV \cite{2020An}, which is denoted by the dotted line in Fig. \ref{fig_voronoi}. The Voronoi diagram can be constructed using four algorithms: incremental algorithm, intersect of half planes, divide and conquer algorithm and plane sweep algorithm \cite{Qiang1999Centroidal,4276115,2008Computational}.

Aiming to design the energy-saving path of UAV for data collection, Baek \emph{et al.} \cite{8933037} modify the Voronoi diagram which determines the optimal hovering point of UAV according to the residual energy of sensors, so as to maximize the minimum remaining energy of sensors, extending the network lifetime. Compared with the path planning of UAV based on Voronoi diagram in \cite{2012Cooperative,2012A,2001Coverage}, the total length of flight path of UAV is shortened. However, in practical application scenarios, it is difficult for a single UAV to deal with too many sensors. Wang \emph{et al.} \cite{9120646} study the cooperative data collection system using multiple UAVs, which applied Voronoi diagram to allocate the collection areas of each UAV. Combined with the optimization of the speed, collection position and transmit power of UAV, the trajectories of multiple UAVs are obtained, which effectively shorten the completion time of data collection.

\subsubsection{Probabilistic roadmap}

\begin{figure}[!t]
	\centering
	\includegraphics[width=0.4\textwidth]{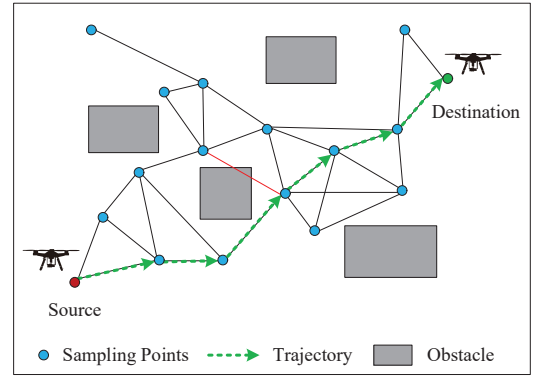}
	\caption{The principle diagram of PRM.}
	\label{fig_PRM}
\end{figure}

PRM consists of sampling points and collision-free straight-line edges \cite{2007On,2003A,2020HPPRM}, as shown in Fig. \ref{fig_PRM}. The sampling points are selected in the space without obstacles. PRM based path searching algorithm consists of the stages of learning and query. In the learning stage, PRM randomly scatters points in the region to construct an undirected graph. In the query stage, PRM applies the shortest path algorithms to find the flight path of UAV. This method uses relatively few random sampling points to find a solution. And the probability of finding a path tends to 1 with the increase of the number of sampling points.

\begin{figure}[htbp]
	\centering
	\subfigure[{The first step.}]{
		\includegraphics[width=3.3cm]{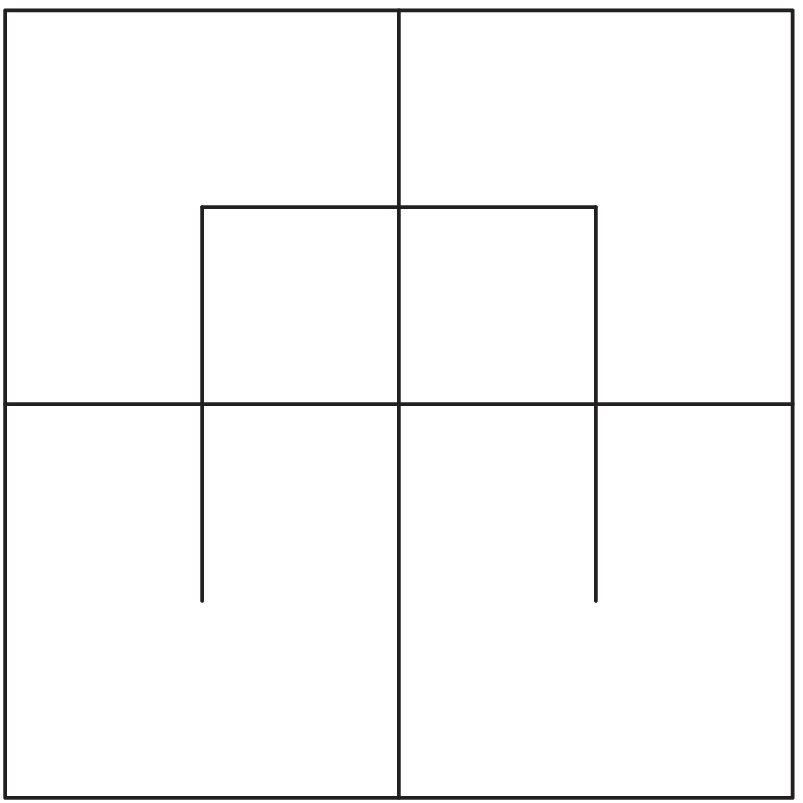}
	}
	\quad
	\subfigure[{The second step.}]{
		\includegraphics[width=3.3cm]{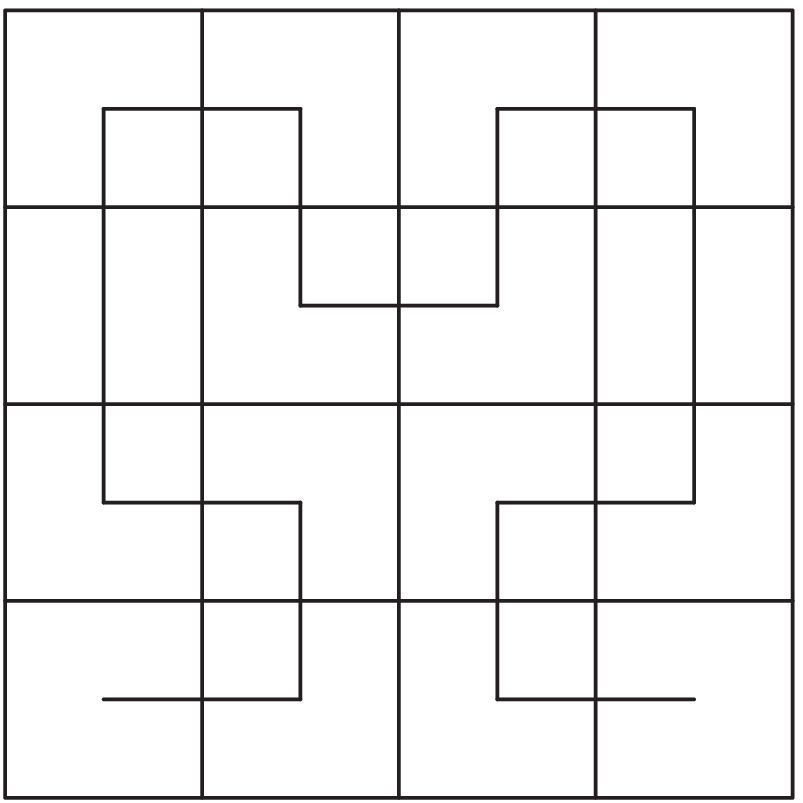}
	}
	\quad
	\subfigure[{The third step.}]{
		\includegraphics[width=3.3cm]{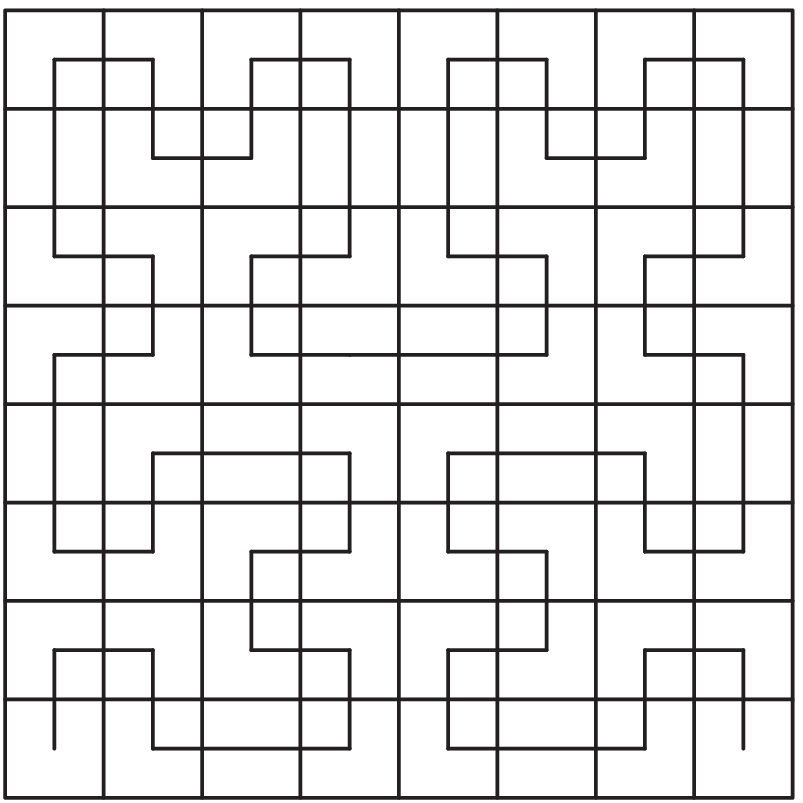}
	}
	\quad
	\subfigure[The fourth step.]{
		\includegraphics[width=3.3cm]{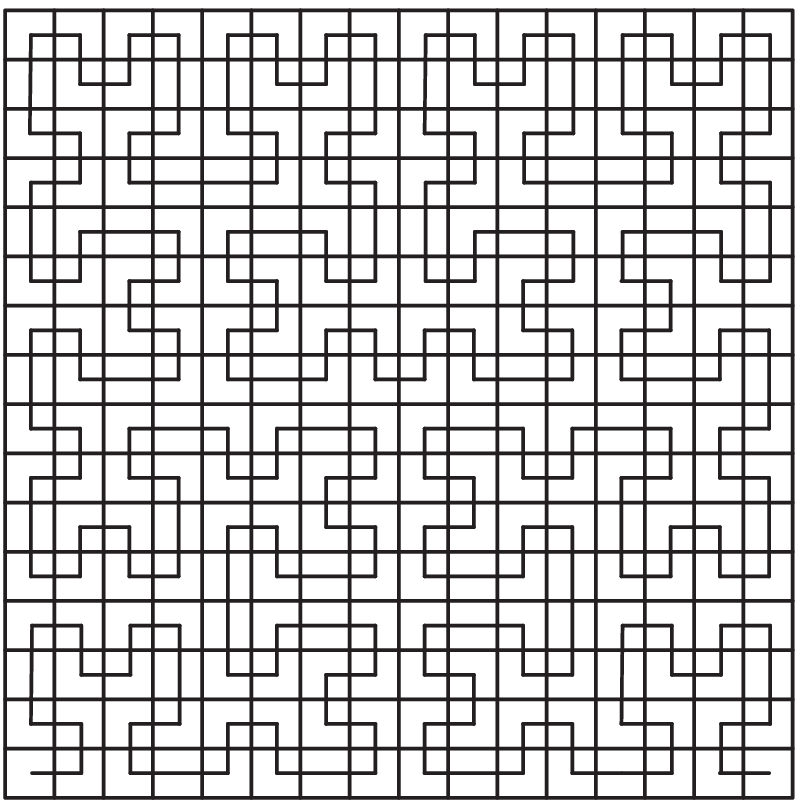}
	}
	\caption{The first three steps of the Hilbert space-filling curve.}
	\label{fig_Hilbert}
\end{figure}

\begin{table*}[!t]
	\label{Table_5}
	\small
	\caption{Comparison of optimization theory based algorithms}
	\renewcommand{\arraystretch}{1.1} 
	\begin{center}
		\begin{threeparttable}
			\begin{tabular}{|m{0.1\textwidth}<{\centering}|m{0.08\textwidth}<{\centering}|m{0.2\textwidth}<{\centering}|m{0.22\textwidth}<{\centering}|m{0.05\textwidth}<{\centering}|m{0.05\textwidth}<{\centering}|m{0.05\textwidth}<{\centering}|m{0.05\textwidth}<{\centering}|}
				\hline
				Algorithm & Reference & Objective function & Decision variables & AoCD & TE & EE & AoI\\
				\hline
				
				\multirow{6}[57]{\linewidth}{\makecell[c]{Dynamic \\  programming}}
				& \cite{8432487} & Minimization of flight time of UAV. & Time slot, speed of UAV, transmit power of sensors. & $\surd$ & $\surd$ & $\surd$ & $\times$ \\
				\cline{2-8}
				& \cite{8377357} & Minimization of flight time of UAV. & Time slot, transmit power of sensors, data collection interval. & $\surd$ & $\surd$ & $\surd$ & $\times$ \\
				\cline{2-8}
				& \cite{8756665} & Minimization of AoI. & The number and location of data collection points, UAV path planning, associations of sensors and data collection points. & $\times$ & $\surd$ & $\times$ & $\surd$ \\
				\cline{2-8}
				& \cite{8756751} & Minimization of AoI. & Data collection mode, energy consumption of sensors, age evolution of collected data. & $\times$ & $\times$ & $\surd$ & $\surd$ \\
				\cline{2-8}
				& \cite{9151993} & Minimization of the average AoI. & Energy harvesting time, data collection time for each
				sensor, UAV path planning. & $\times$ & $\times$ & $\times$ & $\surd$ \\
				\cline{2-8}
				& \cite{9286911} & Minimization of AoI. & The number of data collection points, transmission priority of sensors, the balance between flight time of UAV and data upload time, UAV path planning. & $\times$ & $\surd$ & $\times$ & $\surd$ \\
				\cline{2-8}
				\hline
				
				\multirow{1}[1]{\linewidth}{\makecell[c]{Branch and \\ bound}}
				& \cite{8842600} & Maximization of the number of sensors and minimization of the amount of collected data with time constraint. & Radio resource allocation, UAV path planning. & $\surd$ & $\times$ & $\times$ & $\surd$ \\
				\cline{2-8}
				
				\hline
			\end{tabular}
			\begin{tablenotes}
				\footnotesize
				\item[1] $\surd$ denotes the existence of the feature, $\times$ denotes the absence of the feature
				\item[2] AoCD: Amount of collected data, TE: Time efficiency, EE: Energy efficiency, AoI: Age of information
			\end{tablenotes}
		\end{threeparttable}
	\end{center}
\end{table*}

Penicka \emph{et al.} \cite{8741059} consider the existence of obstacles in UAV assisted data collection. The data collection mission is considered as a physical orienteering problem, which aims at ensuring a feasible, collision-free trajectory to maximize the collected data (reward) with the constraint of energy (budget). The asymptotically optimal sampling-based probabilistic roadmap (PRM*) combined with variable neighborhood search (VNS) algorithm is proposed. The proposed algorithm applies VNS algorithm to expand the initial roadmap with low-density nodes, and the PRM* algorithm is applied to build a shortened collision-free trajectory. The low-density initial roadmap brought an advantage of lower computational complexity compared with other algorithms with high-density initial roadmaps.

\subsubsection{Hilbert curve}

The Hilbert curve plays an important role in image processing, multidimensional data index, and so on, which is a space-filling curve mapping from one-dimensional space to the two-dimensional space \cite{2004Four}. As shown in Fig. \ref{fig_Hilbert}, a space is divided into subspaces, and the Hilbert curve can pass through all the subspaces. Fig. \ref{fig_Hilbert} shows the Hilbert curve from first order to fourth order.

The construction of Hilbert curve is an iteration of the order. The curve is organized by straight lines, and always pass through the center of subspaces. Each subspace is divided into four subspaces in each iteration, such that the curve can have better granularity after iteration. Finally, a curve covering the entire space is formed, which provides the optimal trajectory for the UAV to cover the area of data collection.
\begin{figure}[!t]
	\centering
	\includegraphics[width=0.4\textwidth]{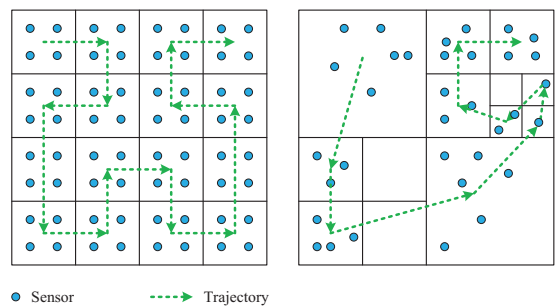}
	\caption{Trajectory of UAV based on Hilbert curve.}
	\label{fig_Hilbert_new}
\end{figure}
In \cite{8943139}, a UAV using delay tolerant network protocol is applied to realize data collection in the large-area IoT scenarios without pre-deployed network infrastructure. In particular, Liang \emph{et al.} \cite{8943139} apply the path planning algorithm based on Hilbert curve to determine the flight path of UAV. The coordinates of sensors are applied as one of the inputs of the algorithm. As shown in Fig. \ref{fig_Hilbert_new}, the trajectory generated by the algorithm based on Hilbert curve can traverse all the cells where sensors are deployed and skip the cells without sensors.

\subsection{Optimization Theory based Algorithms }

\begin{table*}[!t]
	\label{Table_5}
	\small
	\caption{Comparison of optimization theory based algorithms}
	\renewcommand{\arraystretch}{1.1} 
	\begin{center}
	\begin{threeparttable}	
		\begin{tabular}{|m{0.11\textwidth}<{\centering}|m{0.07\textwidth}<{\centering}|m{0.2\textwidth}<{\centering}|m{0.22\textwidth}<{\centering}|m{0.05\textwidth}<{\centering}|m{0.05\textwidth}<{\centering}|m{0.05\textwidth}<{\centering}|m{0.05\textwidth}<{\centering}|}
			\hline
			Algorithm & Reference & Objective function & Decision variables & AoCD & TE & EE & AoI\\
			\hline
			
			\multirow{9}[50]{\linewidth}{\makecell[c]{Successive \\ convex \\ approximation}}
			& \cite{8757098} & Minimization of propulsion energy consumption of UAV. & Length and curvature of UAV path. & $\times$ & $\times$ & $\surd$ & $\times$ \\
			\cline{2-8}
			& \cite{8714077} & Minimum-maximum energy consumption of UAV and sensors. & Communication scheduling between UAV and sensors, transmit power of sensors, UAV path planning. & $\times$ & $\times$ & $\surd$ & $\times$ \\
			\cline{2-8}
			& \cite{9013148} & Weighted minimization of energy consumption for UAV and sensors. & UAV path planning, data collection time, wake-up scheduling of sensors. & $\times$ & $\surd$ & $\surd$ & $\times$ \\
			\cline{2-8}
			& \cite{8779596} & Minimization of data collection time. & Wake-up scheduling of sensors, UAV path planning. & $\surd$ & $\surd$ & $\surd$ & $\times$ \\
			\cline{2-8}
			& \cite{9273074} & Maximum of minimum data collection rate. & UAV time allocation, UAV path planning. & $\surd$ & $\surd$ & $\surd$ & $\times$ \\
			\cline{2-8}
			& \cite{9174765} & Maximization of throughput between UAV and sensors. & Resource allocation, UAV path planning. & $\times$ & $\times$ & $\times$ & $\times$ \\
			\cline{2-8}
			& \cite{8877250} & Minimization of flight time of UAV. & Resource allocation of sensors, UAV path planning. & $\surd$ & $\surd$ & $\surd$ & $\times$ \\
			\cline{2-8}
			& \cite{8887242} & Maximization of power transmission efficiency. & Power of the charging station, UAV path planning, communication scheduling between of UAV and sensors. & $\times$ & $\times$ & $\surd$ & $\times$ \\
			\cline{2-8}
			& \cite{9121338} & Maximization of throughput of air-to-ground networks. & Communication scheduling between of UAV and sensors, UAV path planning, transmit power of sensor/UAV-CHs. & $\times$ & $\times$ & $\surd$ & $\times$ \\
			\cline{2-8}
			\hline
			
			\multirow{1}[1]{\linewidth}{\makecell[c]{Matrix \\ completion  }}
			& \cite{9171476} & Minimization of energy consumption of UAV and redundant data. & Sampling point selection, UAV path planning. & $\surd$ & $\times$ & $\surd$ & $\times$ \\
			\cline{2-8}
			
			\hline
		\end{tabular}
	\begin{tablenotes}
		\footnotesize
		\item[1] $\surd$ denotes the existence of the feature, $\times$ denotes the absence of the feature
		\item[2] AoCD: Amount of collected data, TE: Time efficiency, EE: Energy efficiency, AoI: Age of information
	\end{tablenotes}
	\end{threeparttable}

	\end{center}
\end{table*}

In the problem of joint path planning and resource allocation of UAV for data collection, the algorithms based on optimization theory, such as dynamic programming (DP), branch and bound, successive convex approximation (SCA), can find the optimal or suboptimal solution. However, for some NP-hard problems, the time complexity is exponential and intolerable. When the decision space is huge, the heuristic methods can be applied to find the feasible solution. Table \uppercase\expandafter{\romannumeral6} and Table \uppercase\expandafter{\romannumeral7} compare the literatures using optimization theory based algorithms.

\subsubsection{Dynamic programming}

DP is a nonlinear optimization method proposed by Bellman \cite{PUTERMAN2003673}. DP transforms multi-stage decision-making problem into a series of single-stage decision-making problems \cite{9249903}. The optimal path can be obtained recursively. According to the types of objective functions, the DP based joint path planning and resource allocation methods are summarized as follows.

$\bullet$ \emph{Minimization of flight time of UAV}

Minimizing the flight time of UAV can fundamentally reduce the energy consumption of UAV. The sensors are deployed in a straight line in [35] and the trajectory of UAV is divided into non-overlapping intervals for data collection. The interval for data collection is optimized using DP to minimize the flight time of UAV. And the optimal speed of UAV and transmit power of sensors are found. Furthermore, Gong \emph{et al.} \cite{8377357} apply DP to optimize the partition of intervals. 

$\bullet$ \emph{Minimization of AoI}

In terms of the average AoI, taking AoI as objective function is better than taking flight time of UAV as optimization objective.  Especially when the sensor buffer is limited or data is covered regularly, timely data collection becomes very critical. Jia \emph{et al.} \cite{8756751} find the best path for UAV by jointly considering the access sequence of sensors and data collection mode. A solution based on DP is proposed to determine the optimal access sequence of sensors and minimize the average AoI. Hu \emph{et al.} \cite{9151993} decompose the problem of minimizing the average AoI of collected data into time allocation and energy transfer problem,  and trajectory of UAV optimization problem. Different from other models considering that UAV has limited energy and needs to be powered by a BS, Hu \emph{et al.} \cite{9151993} assume that UAV acting as a mobile power source collects the data of sensors and powers the sensors. However, the algorithms using DP are complex and inefficient when the number of sensors is large.

\begin{figure*}
	\centering
	\subfigure[{The schematic diagram of SCA.}]{
		\label{fig_SCA}
		\includegraphics[width=0.35\textwidth]{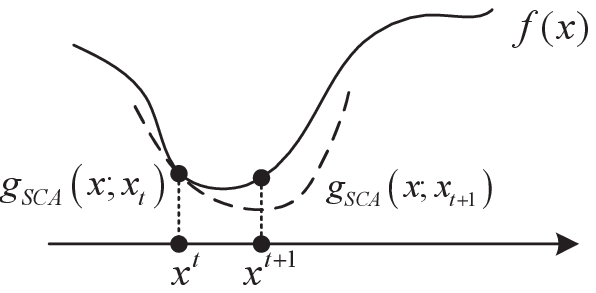}}
	\hspace{0.5cm}
	\subfigure[{The schematic diagram of MM.}]{
		\label{fig_MM}
		\includegraphics[width=0.35\textwidth]{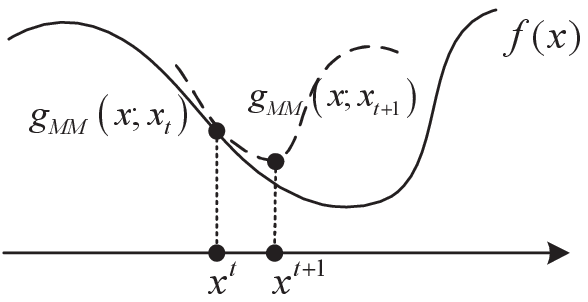}}
	\caption{Comparison of MM and SCA for solving optimization problems.}
	\label{fig_SCA_MM}
\end{figure*}

$\bullet$ \emph{Minimization of maximum AoI or average AoI}

Aiming at the problem of data collection with optimal AoI, sensor association and path planning are jointly optimized to reveal the optimal tradeoff between data transmission time of sensors and the flight time of UAV in [9]. It is verified that the path maximizing AoI is the shortest Hamiltonian path. When the number of data collection points, flight path of UAV and sensors scheduling are jointly optimized, maximum AoI can be greatly reduced. Liu \emph{et al.} \cite{9286911} study the problem of UAV assisted data collection based on AoI. On the basis of determining the hover point and sensors access sequence of UAV, they use DP to calculate optimal trajectories to minimize the maximum AoI or average AoI.

\subsubsection{Branch and bound}

Branch and bound (BB) is a kind of search and iteration method \cite{2002Branch,6224902}. BB decomposes the given problem into sub-problems iteratively. The sub-problem decomposing process is called branching. The boundary of the sub-problem's objective function is called the bound.

In order to collect data from as many sensors as possible and ensure a minimum amount of uploaded data per sensor, Samir \emph{et al.} \cite{8842600} design a high-complexity branch, reduce and bound (BRB) algorithm to find the global optimal solution in relatively small scale IoT. In the BRB algorithm, a set of $N$ non-overlapping hyper-rectangles that cover the optimization problem of maximizing the number of served sensors is maintained. The hyper-rectangle contains all feasible solutions of the optimization model. And three operations, namely, branching, reduction and bounding, are conducted for each iteration in the BRB to improve the lower and upper bounds of the objective function  until the difference between the lower and upper bounds is smaller than a predefined value. Accordingly, the hyper-rectangle is constantly reduced to update the upper and lower bounds and remove the part that does not meet the feasible solution.

\subsubsection{Successive convex approximation}

SCA transforms the original nonconvex optimization problem into a series of convex optimization problems which can be solved efficiently. As shown in Fig. \ref{fig_SCA_MM}, the principle of SCA is similar to majorize-minimize or minorize-maximize (MM), which can solve a series of convex optimization problems similar to the original problem iteratively \cite{7547360,6363727}. When the final convergence condition is satisfied, the solution is approximately regarded as the solution of the original problem. MM requires that the approximation function $U_{MM}(x_{t})$ is the upper bound of the original function at the approximation point, which is one feasible solution of objective function. Besides, SCA requires that the approximation function $U_{SCA}(x_{t})$ is convex. Then, we summarize the SCA based joint path planning and resource allocation methods according to the objective functions.

$\bullet$ \emph{Minimization of energy consumption}

The path planning of UAV needs to take into account time and energy consumption. Dong \emph{et al.} \cite{8757098} firstly propose a mobility model for fixed wing UAV, which is composed of straight line segment and circle segment. And they apply path discretization approach and SCA to optimize the straight circular trajectory to minimize the propulsion energy consumption of UAV on the premise of satisfying the throughput requirements. Zhan and Lai \cite{8714077} focus on the propulsion energy model of rotary wing UAV. Under the premise of limited UAV energy, they jointly optimize the communication scheduling of sensors, transmit power allocation and trajectory of UAV to minimize the maximum energy consumption of UAV and sensors. For the non-convex problem, an efficient suboptimal solution by applying the alternating optimization and SCA is proposed. In particular, Zhan and Huang \cite{9013148} reveal a basic trade-off between the energy consumption of UAV and sensors. In \cite{9013148}, they optimize the trajectory of UAV, data collection time and wake-up schedule to minimize the weighted energy consumption of UAV and sensors. In addition, SCA and block coordinate descent (BCD) techniques are applied to obtain the local optimal solution.

$\bullet$ \emph{Maximization of flight time of UAV}

Zhan and Zeng \cite{8779596} minimize the completion time of the maximum task of multiple UAVs and ensure that the energy of all sensors are sufficient to upload data. By levering bisection method and time discretization technique, the original problem is transformed into a discrete equivalent problem, which contains a finite number of optimization variables. And a Karush-Kuhn-Tucker (KKT) solution is obtained by using SCA. In \cite{8877250}, two orthogonal multiple access schemes, namely, TDMA and frequency division multiple access (FDMA), are designed and compared. Zong \emph{et al.} use a binary search algorithm based on SCA to minimize the completion time of data collection. It is proved that the completion time in FDMA scheme is shorter than that in TDMA scheme.

$\bullet$ \emph{Maximization of the amount of collected data}

In \cite{9273074}, Luo \emph{et al.} propose an efficient iterative algorithm to optimize the 3D path planning and time allocation for UAV. The algorithm jointly utilizes epigraph equivalent representation, variable substitution, equivalent representation by using SCA. In addition, the combination of wireless power transmission and data collection effectively solves the problem of data collection with limited energy. In \cite{9055113}, considering the velocity constraint of UAV, the corresponding transmission scheduling and power allocation for sensors are designed by applying BCD and SCA under the given initial trajectory.

\begin{figure*}[!t]
	\centering
	\includegraphics[width=0.9\textwidth]{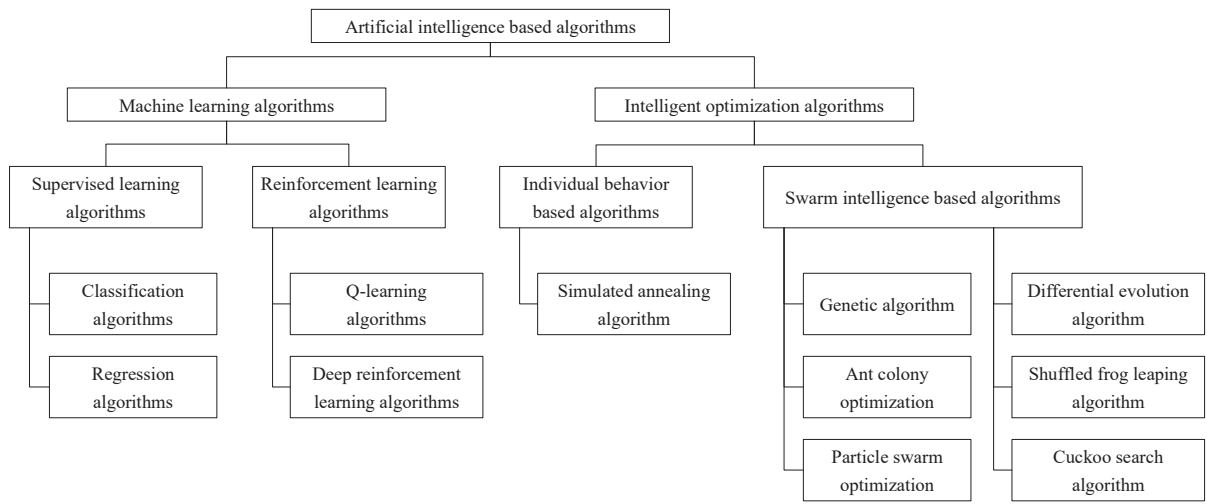}
	\caption{Artificial intelligence based algorithms for joint path planing and resource allocation.}
	\label{fig_AIclassification}
\end{figure*}

$\bullet$ \emph{Maximization of throughput}

With the goal of maximization of average throughput between UAV and  sensors, Sun \emph{et al.} \cite{9174765} propose a two-tier UAV communication strategy. In the first tier of the strategy, sensors transmit data to their CHs via a multi-channel ALOHA-based random access scheme. And in the second tier, CHs deliver the aggregated data to the UAV through coordinated TDMA. A low complexity iterative algorithm based on SCA is designed to jointly optimize the trajectory and resource allocation of UAV. Under the constraints of UAV mobility and transmit power of sensors or UAV, Hua \emph{et al.}  \cite{9121338} combine transmit power and communication scheduling of sensor or UAV to maximize the total throughput of the system. And they use SCA to obtain the global optimal solution and suboptimal solution maximizing throughput, respectively.

$\bullet$ \emph{Maximization of power transmission efficiency}

Chen \emph{et al.}  \cite{8887242} introduce the adaptive resonant beam charging system to charge the UAV. In addition to ensuring the quality of service, the power transmission efficiency is maximized by jointly optimizing the trajectory of UAV and the power of charging station. In particular, SCA and Dinkelbach methods are used to deal with the problems of trajectory design and power control, respectively.

\subsubsection{Matrix completion method}

Matrix completion uses the observed matrix elements to estimate the unknown elements and recover the entire matrix \cite{9068243,7295550}. These data matrices are usually low rank, and there are some missing data. The problem of low rank matrix completion is to predict those missing data through the observed data and then recover the matrix, using the low rank property of the matrix. Based on the above characteristics, matrix completion method can be used to select the UAV data sampling points and recover the missing data. Because of the high correlation between the collected data, even if the UAV only collects limited amount of data, it can recover the data of the entire monitoring area by exploring the data correlation \cite{2017Bloom,8108213}. 

Matrix completion method is applied to guide UAV to select data collection points from the perspective of time and space in \cite{9171476}. The data collected at different positions/times can form a position/time matrix. Taking the location matrix as an example, according to the number of selected sampling points in the row of the sampling points in the location matrix, \cite{9171476} dynamically adjust the probability of the selected sampling points, which effectively reduces the data redundancy. Based on the results of matrix completion method, ant colony
optimization (ACO) is used to minimize the trajectory of UAV. According to the path selection probability, the sampling points are gradually selected by UAV until their number reaches a certain value.

\subsection{Artificial Intelligence based Algorithms }

Compared with the traditional algorithms, AI based algorithms can deal with the uncertainty of path planning more effectively and get the approximately optimal solution. AI based the path planning algorithms are classified into machine learning algorithms and intelligent optimization algorithms. The detailed classification is shown in Fig. \ref{fig_AIclassification}.

\subsubsection{Supervised learning}

A machine learning algorithm is able to acquire knowledge autonomously. According to learning methods, machine learning algorithms can be classified into semi-supervised learning, unsupervised learning, supervised learning, ensemble learning, reinforcement learning (RL) and deep learning (DL). Among them, supervised learning and RL are widely used in UAV path planning. Table \uppercase\expandafter{\romannumeral8} compares the literatures using supervised learning based algorithms.

\paragraph{Classification algorithms}

The classification algorithms mainly construct the classification model in the labeled training data, and then classify the new data based on this model. In \cite{8913466}, the path planning problem is modeled as a traveling salesman problem (TSP) with the neighborhood (TSPN). And two trajectory design methods, segment-based trajectory optimization algorithm (STOA) and group-based trajectory optimization algorithm (GTOA), are proposed. STOA calculates the visit order of all sensors and data collection points. While GTOA divides all sensors into clusters and calculates the visit order and locations based on clusters. UAV only needs to pass through the cross region of common transmission region of the grouped sensors. In particular, the sensor that has the largest number of neighbor sensors is selected as CH. The limited flight time of UAV is a challenge for data collection. Aiming to minimizing the total energy consumption of sensors while ensuring data collection through joint optimization of trajectory of UAV, data transmission scheduling of sensors and transmit power. Zhao \emph{et al.} \cite{9175017} innovatively use non-orthogonal multiple-access (NOMA) in UAV data collection. In addition, the generalized benders decomposition (GBD) is used to decouple sensers scheduling and transmit power.

\begin{table*}[!t]
	\label{Table_5}
	\small
	\caption{Comparison of supervised learning based algorithms}
	\renewcommand{\arraystretch}{1.1} 
	\begin{center}
		\begin{threeparttable}	
			\begin{tabular}{|m{0.1\textwidth}<{\centering}|m{0.08\textwidth}<{\centering}|m{0.2\textwidth}<{\centering}|m{0.22\textwidth}<{\centering}|m{0.05\textwidth}<{\centering}|m{0.05\textwidth}<{\centering}|m{0.05\textwidth}<{\centering}|m{0.05\textwidth}<{\centering}|}
				\hline
				Algorithm & Reference & Objective function & Decision variables & AoCD & TE & EE & AoI\\
				\hline

				\multirow{2}[10]{\linewidth}{\makecell[c]{Classification \\ algorithm}}
				& \cite{8913466} & Minimization of flight time of UAV. & Height and speed of UAV, link scheduling, UAV path planning. & $\times$ & $\surd$ & $\times$ & $\times$ \\
				\cline{2-8}
				& \cite{9175017} & Minimization of total energy consumption of sensors. &  Transmission scheduling of sensors, transmit power of sensors, UAV path planning. & $\surd$ & $\times$ & $\surd$ & $\times$ \\
				\cline{2-8}
				\hline
				
				\multirow{1}[1]{\linewidth}{\makecell[c]{Regression \\ algorithm}}
				& \cite{8927929} & Maximization of the amount of data collected. & Clustering of sensors, value of information (VoI), power of sensors, UAV path planning. & $\surd$ & $\times$ & $\surd$ & $\times$ \\
				\cline{2-8}
				
				\hline
			\end{tabular}
			\begin{tablenotes}
				\footnotesize
				\item[1] $\surd$ denotes the existence of the feature, $\times$ denotes the absence of the feature
				\item[2] AoCD: Amount of collected data, TE: Time efficiency, EE: Energy efficiency, AoI: Age of information
			\end{tablenotes}
		\end{threeparttable}
	\end{center}
\end{table*}

\paragraph{Regression algorithms}

\begin{table*}[!t]
	\label{Table_5}
	\small
	\caption{Comparison of Q-learning based algorithms}
	\renewcommand{\arraystretch}{1.1} 
	\begin{center}
	\begin{threeparttable}	
		\begin{tabular}{|m{0.1\textwidth}<{\centering}|m{0.08\textwidth}<{\centering}|m{0.2\textwidth}<{\centering}|m{0.22\textwidth}<{\centering}|m{0.05\textwidth}<{\centering}|m{0.05\textwidth}<{\centering}|m{0.05\textwidth}<{\centering}|m{0.05\textwidth}<{\centering}|}
			\hline
			Algorithm & Reference & Objective function & Decision variables & AoCD & TE & EE & AoI\\
			\hline

			\multirow{5}[27]{\linewidth}{\makecell[c]{Q-learning}}
			& \cite{9114970} & Maximization of the amount of data collected. & Time slot, UAV path planning. & $\surd$ & $\surd$ & $\times$ & $\times$ \\
			\cline{2-8}
			& \cite{9075299} & Asymptotic minimization of packet loss in IoT. & scheduling of microwave power transfer (MPT) of UAV, UAV path planning. & $\surd$ & $\times$ & $\surd$ & $\times$ \\
			\cline{2-8}
			& \cite{8752017} & Minimization of data expiration and loss. & UAV path planning. & $\surd$ & $\times$ & $\times$ & $\surd$ \\
			\cline{2-8}
			& \cite{9121767} & Minimization of flight path length of UAV. & UAV collision avoidance, UAV path planning. & $\times$ & $\surd$ & $\times$ & $\times$ \\
			\cline{2-8}
			& \cite{9217347} & Minimization of the total energy consumption of UAV. & The real-time amount of data collected, the transmission time between the UAV and aerial BS, UAV path planning. & $\surd$ & $\surd$ & $\surd$ & $\times$ \\
			\cline{2-8}
			\hline
			
		\end{tabular}
	\begin{tablenotes}
		\footnotesize
		\item[1] $\surd$ denotes the existence of the feature, $\times$ denotes the absence of the feature
		\item[2] AoCD: Amount of collected data, TE: Time efficiency, EE: Energy efficiency, AoI: Age of information
	\end{tablenotes}
		\end{threeparttable}

	\end{center}
\end{table*}

\begin{table*}[!t]
	\label{Table_5}
	\small
	\caption{Comparison of deep reinforcement learning based algorithms}
	\renewcommand{\arraystretch}{1.1} 
	\begin{center}
		\begin{threeparttable}
			
			\begin{tabular}{|m{0.11\textwidth}<{\centering}|m{0.08\textwidth}<{\centering}|m{0.2\textwidth}<{\centering}|m{0.21\textwidth}<{\centering}|m{0.05\textwidth}<{\centering}|m{0.05\textwidth}<{\centering}|m{0.05\textwidth}<{\centering}|m{0.05\textwidth}<{\centering}|}
				\hline
				DRL method & Reference & Objective function & Decision variables & AoCD & TE & EE & AoI\\
				\hline

				\multirow{5}[30]{\linewidth}{\makecell[c]{Deep\\ Q-network \\ (Deep \\Q-learning)}}
			
				& \cite{9145249} & Minimization of the weighted sum of AoI, packet drop rate and the energy consumption of UAV. & Random or fixed sampling, buffer management, UAV path planning. & $\surd$ & $\times$ & $\surd$ & $\surd$ \\
				\cline{2-8}
				& \cite{9162896} & Minimization of weighted sum of AoI. & Transmission scheduling of sensors, UAV path planning. & $\times$ & $\surd$ & $\surd$ & $\surd$ \\
				\cline{2-8}
				& \cite{9148316} & Minimization of buffer overflow and transmission failure of sensors. & Instantaneous waypoint of UAV. & $\surd$ & $\times$ & $\times$ & $\times$ \\
				\cline{2-8}
				& \cite{9528844} & Minimization of the average AoI. & Transmission scheduling and energy harvesting of sensors, UAV path planning. & $\times$ & $\times$ & $\surd$ & $\surd$ \\
				\cline{2-8}
				& \cite{9701330} & Minimization of flight path length of UAV and maximization of the amount of data collected. & UAV path planning. & $\surd$ & $\surd$ & $\times$ & $\times$ \\
				\cline{2-8}
				\hline
			
				\multirow{5}[30]{\linewidth}{\makecell[c]{Deep \\ deterministic \\ strategy \\ gradient}}
				
				& \cite{9195789} & Minimization of AoI under minimum throughput constraints. & Scheduling strategy of UAV, UAV path planning. & $\times$ & $\times$ & $\times$ & $\surd$ \\
				\cline{2-8}
				& \cite{9117104} & Minimization of flight time of UAV. & Time window of data collection, priority of sensors, UAV path planning. & $\times$ & $\surd$ & $\surd$ & $\times$ \\
				\cline{2-8}
				& \cite{9238897} & Long term freshness of UAV situation awareness. & Resource allocation, UAV path planning. & $\times$ & $\times$ & $\surd$ & $\surd$ \\
				\cline{2-8}
				& \cite{9426899} &  Minimization of the weighted sum of expected AoI, propulsion energy consumption of UAV and energy consumption of sensors. & Hovering points and flight speed of UAV, transmission scheduling of sensors. & $\times$ & $\times$ & $\surd$ & $\surd$ \\
				\cline{2-8}
				\hline
				
				\multirow{1}[1]{\linewidth}{\makecell[c]{Proximal \\policy \\ optimization }}
				
				& \cite{9155535} &  UAV crowdsensing. & The number of UAV and charging station, UAV path planning. & $\times$ & $\times$ & $\surd$ & $\times$ \\
				\cline{2-8}
				
				\hline
			\end{tabular}
			\begin{tablenotes}
				\footnotesize
				\item[1] $\surd$ denotes the existence of the feature, $\times$ denotes the absence of the feature
				\item[2] AoCD: Amount of collected data, TE: Time efficiency, EE: Energy efficiency, AoI: Age of information
			\end{tablenotes}
		\end{threeparttable}
		
	\end{center}
\end{table*}

The regression algorithms extract features from the sample data and predict the continuous target values corresponding to the new data. Chen \emph{et al.} \cite{8927929} introduce the concept of measurement stream to transform the traditional RL problem into a supervision learning problem. On the premise of defining VoI, \cite{8927929} divides all sensors into clusters, and determines the CHs and data forwarding rules according to the power of sensors and VoI. Finally, the UAV collects the aggregated data from CHs. And direct future prediction (DFP) model is applied in the path planning of UAV, so that UAV can deal with multi-objective tasks, maximizing the VoI of collected data and ensuring the UAV charging.

\subsubsection{Reinforcement learning algorithms}

RL applies the interaction with the environment to explore the mapping between the optimal state and the action, and finally achieves an optimal strategy and maximizes the cumulative revenue. The training data of supervised learning is generally independent with each other. However, RL deals with sequential decision-making problems, which are dependent with each other in the sequencing. RL is usually applied in obstacle avoidance and path planning of UAV. 

\paragraph{Q-learning algorithms}

Q-learning is a model-free RL algorithm using a time-series difference method, which can carry out off-policy learning \cite{5642228}. Q-learning selects new actions and updates value function. Table \uppercase\expandafter{\romannumeral9} compares the
literatures using Q-learning algorithms. We summarize Q-learning based methods according to the optimization objectives.

$\bullet$ \emph{Maximization of the amount of collected data and minimization of data loss rate}

On the premise that the sensor position is unknown, Cui \emph{et al.} model UAV path planning as a Markov decision process (MDP) in \cite{9114970}, as well as divide the target area into multiple identical rectangular cells. \cite{9114970} proposes an RL problem to maximize the cumulative amount of data collected and find the optimal trajectory of UAV. Two trajectory optimization algorithms are proposed based on state-action-reward-state-action and Q-learning so that the flight path of UAV can be optimized without the information on the network's topology. However, \cite{9114970} does not consider energy consumption of UAV and sensors and the packet loss of sensors. In this regard, Li \emph{et al.} \cite{9075299} use UAV with microwave power transfer (MPT) capability and propose a double-Q learning scheduling algorithm, which jointly optimizes the schedule of MPT and data collection  without prior information on battery levels and data queue lengths of sensors. This algorithm effectively minimizes the packet loss of sensors in a long time. Facing the requirement of time-sensitive data collection, Li \emph{et al.} \cite{8752017} propose an RL method to obtain a minimum-maximum AoI optimal path. And they combine AoI, the deadline constraints of data and Q-learning to optimize the flight path of UAV for the first time. The data loss rate is closely related to the flight time of UAV and the flight order over sensors. The results show that the optimization method based on RL has low time consumption and data loss rate.

$\bullet$ \emph{UAV collision avoidance and minimization of energy consumption}

Facing the possible collision of multiple UAVs in data collection, Hsu and Gau \cite{9121767} focus on the collision avoidance algorithm based on RL to obtain an optimal trajectory of UAV. Considering the influence of limited energy and communication resources on flight endurance and data collection of UAV, Ni \emph{et al.} \cite{9217347} use an RL algorithm to optimize the real-time field data size collected by the on-board
camera, transmission time between UAV and aerial BS,  flight path of UAV, so as to minimize the total energy consumption of UAV during data collection.

\paragraph{Deep reinforcement learning algorithms}

DL has strong perception, but it lacks decision-making ability. RL has the ability of decision-making, but it lacks perceptual ability. The combination of them, namely, the deep reinforcement learning (DRL), can complement each other and decide the action to maximize revenue \cite{9321625}. Using DPL to solve the joint optimization problem is also the cutting-edge scheme of UAV assisted data collection. We compare the
literatures using DRL algorithms in Table \uppercase\expandafter{\romannumeral10} and summarize them according to the objective functions as follows.

\begin{figure*}[!t]
	\centering
	\includegraphics[width=0.55\textwidth]{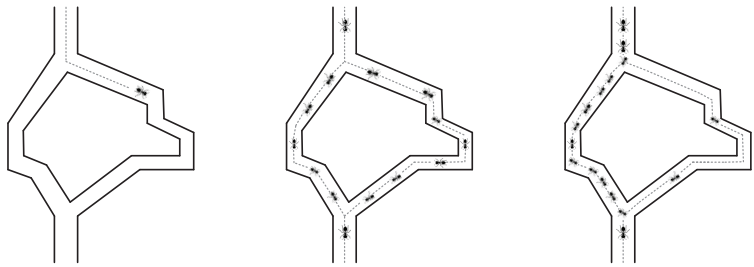}
	\caption{Illustration of ant colony seeking optimal path in foraging.}
	\label{fig_ACO}
\end{figure*}

$\bullet$ \emph{Minimization of AoI} 

Modeling DPL problem as MDP is a common method for simplification. \cite{9145249,9162896,9528844} all apply a DRL algorithm called deep Q-network (DQN) to achieve the optimal strategy and effectively overcome the disaster of dimensionality. In \cite{9145249}, the data collection problem is modeled as an MDP with final state and action space. Further, the weighted sum of AoI, packet loss rate and energy consumption of UAV is minimized by optimizing the flight path of UAV. Yi \emph{et al.} \cite{9162896} establish a MDP to find the best flight path of UAV and the transmission scheduling of sensors, so as to minimize the weighted sum of AoI. Liu \emph{et al.} \cite{9528844} establish an MDP with large state spaces to minimize the average AoI, and the flight path of UAV, transmission scheduling and energy harvesting of sensors are jointly optimized. The simulation results show that the increase of flight time of UAV/transmit power or the decrease of the packet size will reduce the average AoI.
	
In \cite{9195789}, the UAV assisted vehicular network is considered, where UAVs are used to collect and process the data of sensors. Samir \emph{et al.} develop the deep deterministic strategy gradient (DDPG) to process the trajectory and the scheduling policy of UAVs to minimize the expected weighted sum of AoI. In order to minimize the weighted sum of AoI and the number of sensors as well as energy consumption of UAV, Sun \emph{et al.} \cite{9426899} comprehensively consider the data collection points, flight speed and bandwidth allocation of UAVs.

$\bullet$ \emph{UAV collision avoidance and minimization of data transmission failure rate}

In order to avoid the possible collisions among UAVs when collecting data from IoT, Liu \emph{et al.} \cite{9155535} transforms the problem of UAV collision avoidance into a partially observable MDP and present a DRL model named ``j-PPO+ConvNTM". The model can make continuous (path planning) and discrete (data collection and charging) decisions for all UAVs. To minimize the buffer overflow probability of sensors and transmission failure due to the path loss of air-ground channel, Li \emph{et al.} \cite{9148316} propose a path planning algorithm named Deep Reinforcement Learning based Trajectory Planning. Considering the battery levels and buffer lengths of sensors, the location of UAV and channel conditions, a DRL algorithm is applied in the path planning of UAV to minimize the data loss.

$\bullet$ \emph{Maximization of the amount of collected data}

Considering the limited energy and flight time of UAV, Nguyen \emph{et al.} \cite{9701330} take minimizing the flight path and maximizing the amount of collected data as the optimization objectives. Through the deep Q-learning and dueling DQL, the UAV can independently decide the next action at each position, so as to obtain a 3D trajectory that can balance the performance of throughput, data collection time and length of flight path. In \cite{9455139}, adopting the full duplex mode, UAV collects data and charges the sensors within its coverage while hovering. A multi-objective optimization problem is proposed, including maximizing the sum data rate and the total harvested energy of sensors as well as minimizing the energy consumption of UAV. The UAV realizes online path planning through a DRL algorithm based on DDPG.

$\bullet$ \emph{Minimization of data collection time}

Bouhamed \emph{et al.} \cite{9117104} use a combination algorithm of DDPG and Q-learning to minimize data collection time. Furthermore, DDPG is used to realize UAV autonomous navigation in the environment with obstacles, while Q-learning is used to determine the task arrangement of UAV to complete data collection as soon as possible.

$\bullet$ \emph{Minimization of energy consumption}

To minimize the energy consumption of UAV, Fan \emph{et al.} \cite{9238897} present a freshness function based on AoI and a DRL algorithm that can solve the continuous online decision-making problem involving multiple UAVs. In the rapidly changing environment, the algorithm can markedly reduce energy consumption and AoI.

\subsubsection{Intelligent optimization algorithms}

Intelligent optimization algorithms are the combination of random search and local search. In the process of intelligent optimization algorithm, learning strategies are used to obtain the information to effectively find the approximately optimal solution. Then, through a comprehensive searching algorithm, the superiority of the solution is improved.

\paragraph{Individual behavior based algorithms}

Individual behavior based algorithms firstly initialize a feasible solution. Then, the algorithm optimize this feasible solution and probabilistically jump out of local optimum to search the near global optimal solution. Taking simulated annealing (SA) algorithm as an example, it is often used to search the approximately global optimal solution in a large solution space. Different from genetic algorithm (GA), simulated annealing algorithm has the feature of probabilistic jumping out of local optimum, which can accelerate path planning. Table \uppercase\expandafter{\romannumeral11} compares the
literatures using intelligent optimization algorithms.

With a goal of improving the energy efficiency of UAV, Liu \emph{et al.} \cite{8620506} propose a path planning scheme of UAV based on matrix completion and SA. Firstly, the sampling points with degrees from high to low are selected as dominator sampling points, virtual dominator sampling points and follower sampling points. And all data in the monitoring area is recovered by matrix completion. Then, based on the number and location of selected sampling points, SA is used to determine the visit order of each sampling point. The proposed scheme can reduce the data redundancy by 50\% and extend the network lifetime of sensors by 17\% compared with the scheme using random sampling point.

\begin{table*}[!t]
	\label{Table_5}
	\small
	\caption{Comparison of  intelligent optimization based algorithms}
	\renewcommand{\arraystretch}{1.2} 
	\begin{center}
		\begin{threeparttable}	
			\begin{tabular}
				{|m{0.11\textwidth}<{\centering}|m{0.07\textwidth}<{\centering}|m{0.2\textwidth}<{\centering}|m{0.22\textwidth}<{\centering}|m{0.05\textwidth}<{\centering}|m{0.05\textwidth}<{\centering}|m{0.05\textwidth}<{\centering}|m{0.05\textwidth}<{\centering}|}
				\hline
				Algorithm & Reference & Objective function & Decision variables & AoCD & TE & EE & AoI\\
				\hline
				
				\multirow{1}[2]{\linewidth}{\makecell[c]{Simulated \\ annealing}}
				& \cite{8620506} & Maximization of UAV energy utilization. & UAV path planning, sampling points. & $\surd$ & $\times$ & $\surd$ & $\times$ \\
				\cline{2-8}
				\hline
				
				\multirow{3}[20]{\linewidth}{\makecell[c]{Genetic \\ algorithm}}
				& \cite{9014306} & Minimization of flight time of UAV. & The number of UAV, the number and location of CHs, UAV path planning. & $\times$ & $\surd$ & $\times$ & $\times$ \\
				\cline{2-8}
				& \cite{9148990} & Minimization of energy consumption of UAV hovering and communication. & MTCDs clustering, UAV hovering point, UAV path planning. & $\times$ & $\times$ & $\surd$ & $\times$ \\
				\cline{2-8}
				& \cite{9045416} & Minimization of energy consumption of UAV. & UAV path planning. & $\times$ & $\surd$ & $\surd$ & $\times$ \\
				\cline{2-8}
				\hline
				
				\multirow{2}[8]{\linewidth}{\makecell[c]{Ant colony \\ optimization}}
				& \cite{9145752} & Maximization of the per-node capacity of UAV. & The number of UAV and service cell, UAV path planning. & $\surd$ & $\times$ & $\times$ & $\times$ \\
				\cline{2-8}
				& \cite{9210225} & Minimization of flight cost of UAV. & Clustering of sensors, UAV path planning. & $\times$ & $\times$ & $\surd$ & $\times$ \\
				\cline{2-8}
				\hline
				
				\multirow{1}[1]{\linewidth}{\makecell[c]{Particle  swarm \\  optimization}}
				& \cite{9145574} & Minimization of energy consumption of UAV. & Battery capacity, network quality of service (QoS), area coverage, UAV path planning. & $\times$ & $\times$ & $\surd$ & $\times$ \\
				\cline{2-8}
				\hline
				
				\multirow{2}[7]{\linewidth}{\makecell[c]{Differential \\ evolution}}
				& \cite{8894454} & Minimum-maximum the energy consumption of UAV, BSs and sensors. & Devices transmission schedule, UAV path planning. & $\times$ & $\surd$ & $\surd$ & $\times$ \\
				\cline{2-8}
				& \cite{9071449} & Minimization of data collection time. & The selection of CHs, UAV path planning. & $\times$ & $\surd$ & $\times$ & $\times$ \\
				\cline{2-8}
				\hline
				
				\multirow{1}[2]{\linewidth}{\makecell[c]{Shuffled frog \\ leaping}}
				& \cite{8691979} & Minimization of energy consumption of UAV. & Data collection time, UAV path planning. & $\times$ & $\times$ & $\surd$ & $\times$ \\
				\cline{2-8}
				\hline\rule{0pt}{15pt}
				
				\multirow{1}[1]{\linewidth}{\makecell[c]{Cuckoo search}}
				& \cite{8644379} & Minimization of energy consumption of UAV and sensors. & UAV path planning, UAV hovering point. & $\times$ & $\times$ & $\surd$ & $\times$ \\
				\cline{2-8}
				\hline
				
			\end{tabular}
			\begin{tablenotes}
				\footnotesize
				\item[1] $\surd$ denotes the existence of the feature, $\times$ denotes the absence of the feature
				\item[2] AoCD: Amount of collected data, TE: Time efficiency, EE: Energy efficiency, AoI: Age of information
			\end{tablenotes}
		\end{threeparttable}
	\end{center}
\end{table*}

\paragraph{Swarm intelligence based algorithms}

Swarm intelligence algorithms consists of GA, ACO, particle swarm optimization (PSO), differential evolution (DE) algorithm, shuffled frog leaping algorithm (SFLA), cuckoo search (CS) algorithm. They applied a large amount of individuals to mimic the solutions. Then, the learning and evolution mechanisms are applied to improve the superiority of solutions. 

$\bullet$ \emph{Genetic algorithm}

GA simulates the process of biological evolution. A solution to the problem is expressed as a chromosome. The algorithm obtains the most suitable chromosome through the operations of population reproduction, crossover, and mutation, so as to obtain the optimal solution.

To solve the path planning problem of UAV, Alfattani \emph{et al.} \cite{9014306} firstly optimize the number and locations of CHs. Then, GA is applied to optimize the number and paths of UAVs to minimize the data collection time. Shen \emph{et al.} \cite{9148990} convert the path planning problem of UAV into a TSP, which is then solved by GA with a goal of minimizing the energy consumption of  UAV and sensors. Joseph \emph{et al.} \cite{9045416} propose GA based path planning of UAV, which minimizes energy consumption of sensors and UAV under the data offloading time windows and communication constraints.

$\bullet$ \emph{Ant colony optimization}

As shown in Fig. \ref{fig_ACO}, ACO simulates the process that ants find the shortest path during foraging. An ant is regarded as a solution of the path planning problem. Ants spread the pheromone during path searching process and the optimal path accumulates the most pheromone \cite{1997Ant}. Thus, the ants are guided to the optimal path, namely, the optimal scheme is searched. In addition, the diversity of ant colony can avoid local optimization.

Aiming at maximizing the per-node capacity of UAV and the amount of collected data, Wei \emph{et al.} \cite{9145752} adjust the number of UAVs and service cells, and design two path planning algorithms. Among them, the ACO based path planning algorithm can minimize the time of data collection time and achieve the per-node capacity closer to its theoretical upper bound. Lin \emph{et al.} \cite{9210225} use precise algorithm and GA to achieve a hierarchical sensor data collection scheme. A path planning algorithm based on ACO is used to improve the scheme to minimize the energy consumption of UAV.

$\bullet$ \emph{Particle swarm optimization}

PSO originated from the study of birds foraging. Using information sharing mechanism, individuals learn from each other to find food faster \cite{2002Particle}. PSO has great advantages in dealing with continuous optimization problems. 

In generally, data retrieving and processing are carried out when UAV completes data collection. In these cases, Shi and Xu \cite{9145574} propose a path planning problem with the constraint called network quality of service (QoS). For the NP hard problem, a PSO based algorithm is used to minimize the flight time of multiple UAVs.

$\bullet$ \emph{Differential evolution algorithm}

Similar to GA, DE is a kind of intelligent optimization search through the cooperation and competition among individuals in the population \cite{5601760}. It has strong global convergence performance and robustness, which is mainly used to solve global optimization problems with continuous variables.

Wang \emph{et al.} \cite{8894454} study data collection and 3D positioning of sensors. As a mobile data collector and aerial anchor node, UAV assists BSs to realize sensor positioning and data collection of sensors far away from BSs. A DE Algorithm is proposed to optimize UAV path and transmission priority of sensors to minimize-maximize the energy consumption of UAV, sensors and BSs. And the Cram\'{e}r-Rao Bound (CRB) is derived to evaluate the performance of the algorithm. The final path in the optimization scheme reduces the energy consumption of UAV by 32\%. With the selection of optimal CHs and clustering of sensors, Chawra and Gupta \cite{9071449} propose a meta-heuristic method based on DE to estimate the delay effective path of each UAV to minimize data collection time. The algorithm consists of four steps including initial population generation, mutation, crossover and selection.

$\bullet$ \emph{Shuffled frog leaping algorithm}

SFLA simulates the information sharing and communication of frogs in foraging. It combines the advantages of the memetic algorithm (MA) and PSO \cite{2017Shuffled}. 

In \cite{8691979}, Kong \emph{et al.} propose an energy efficient algorithm for data collection in sparse sensor network. The improved SFLA is used to further optimize the location and traversal order of data collection points. Then, these orderly data collection points are used to plan the UAV path with minimum energy consumption of UAV. However, the obstacles in the environment are not considered in this paper.

$\bullet$ \emph{Cuckoo search algorithm}

CS algorithm simulates the unique breeding behavior of cuckoo parasitic brood and lévy flight search mechanism to find the optimal solution. It has a better searching performance than GA and PSO algorithm. In the path planning problem, CS can accurately search the global optimal path in a path planning area with fast speed.

Aiming at minimizing the energy consumption of UAV and sensors, Zhu \emph{et al.} \cite{8644379} propose a hybrid hover location selection algorithm based on a non-service-tolerance mechanism, which ensures that a few highly scattered MTCDs are directly served by the BS to reduce the trajectory length of UAV. On this basis, a path planning method based on CS is proposed to jointly optimize the energy consumption and throughput of UAV, the latency of MTCDs and data collection efficiency.

\section{Future Trends}

\begin{figure}[!t]
	
	\centering
	\includegraphics[width=0.45\textwidth]{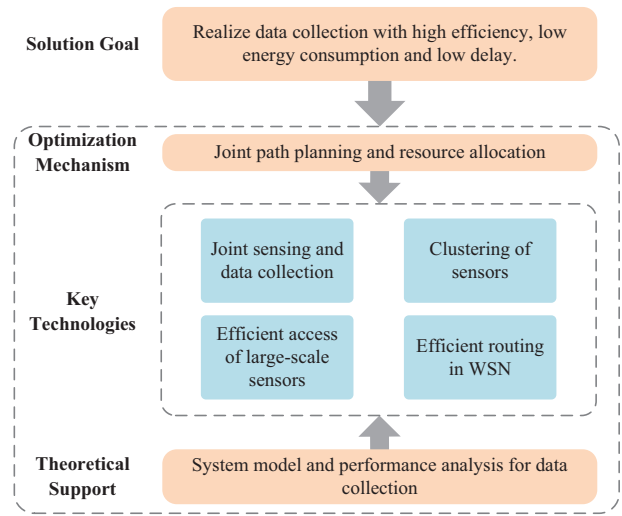}
	\caption{Scheme of UAV assisted data collection.}
	\label{fig_solution}
\end{figure}

It is revealed from the above literature review that most literatures focus on the clustering algorithm of sensors and the path planning algorithm of UAV in order to achieve high efficiency, low energy consumption and low delay data collection. However, the following two challenges have not attracted sufficient attention.
\begin{itemize}
	
	\item In the large-scale IoT, a large amount of sensors transmit data to UAV, which leads to data congestion. 
	\item UAV assisted data collection may be carried out without the knowledge of the locations of sensors.	
	
\end{itemize}

For example, in the scenario of ocean monitoring, there are many buoy sensors widely distributed on the sea surface \cite{9126800}, which may lead to transmission conflicts between competing sensors. In addition, due to insufficient communication coverage of coastal BSs and high positioning costs such as GPS positioning, it is difficult to obtain the locations of sensors in advance. Similar scenarios include agricultural monitoring and soil monitoring \cite{9042335,9265880,8675167}. Therefore, we propose a complete solution for UAV assisted data collection in Fig. \ref{fig_solution}. Joint sensing and data collection, clustering of sensors, efficient access of large-scale sensors and efficient routing in WSN are introduced as key technologies, and joint path planning and resource allocation are applied to provide optimization mechanism with these key technologies.

In the future, we need to focus on but not limited to the following two research trends.

\subsection{Efficient Multiple Access}

MAC protocol is designed to satisfy the performance requirements of the network, including solving the potential transmission conflicts between competing sensors, reducing the average packet delay, increasing network throughput, supporting the access of a large number of sensors, balancing energy consumption of UAV and sensors, and so on. The efficient multiple access between UAV and sensors has a crucial impact on the efficiency of data collection and the freshness of data.

In \cite{8718663}, the performance of MAC protocol is improved by scheduling resources in the dimensions of time, frequency and space. Sotheara \emph{et al.} \cite{7063642} divide the coverage area of UAV into priorities, and sensors access channels according to the priorities of their own regions to reduce transmission conflicts. Due to the limitation of bandwidth, the performance of MAC protocol will reach the bottleneck, especially when the number of sensors is large. The ISM (Industrial, Scientific and Medical) bands including license-free communication frequencies 173 MHz, 433 MHz, 868 MHz and 915 MHz and 2.4 GHz are commonly applied to WSN \cite{gupta2014overview}. In order to expand the bandwidth of WSN, 800/900/1800/2100 MHz of 4G (4th generation mobile communication system) \cite{Rashmi2017A} and 5.15-5.85 GHz band of Wi-Fi \cite{2016Coexistence} are also provided for WSN. The application of mmWave in 5G-A (5G Advanced) and Teraherz (THz) in 6G will solve the problem of spectrum shortage \cite{6162401,5779106}, which has great potential in providing high data rate and low latency for WSN, providing opportunities for the design of efficient MAC protocol.

When the number of sensors is huge, bandwidth and scheduling schemes cannot always satisfy the requirements of fast and efficient access to channels. In this situation, the NOMA technique is promising to further improve the efficiency of MAC protocol \cite{8718663,7972935}. The combination of UAVs and NOMA can definitely provide high spectral efficiency and support massive connectivity in IoT scenarios \cite{8641425}. To the best of our knowledge, the investigation on the application of NOMA in UAV data collection is still in its infancy stage \cite{9126800,9517120}. In order to better exploit both the mobility of the UAV and the efficiency of NOMA to improve the UAV assisted data collection performance in ocean observation networks, Chen \emph{et al.} \cite{9126800} formulated a joint optimization problem of the location of UAV, sensor grouping, and power control in terms of sum rate. When the number of sensors is large, their scheme is able to select sensors with less interference to access the channel, improving the overall performance. Different from \cite{9126800}, the UAV only allows a limited number of nearby devices to be connected to the uplink NOMA network in each time slot, leading to better LoS and successive interference cancelation (SIC) with lower complexity in \cite{9517120}. Mu \emph{et al.} \cite{9454446} deploy an intelligent reflecting surface (IRS) to enhance the transmission from UAVs to their intended users while mitigating the interference caused to other unintended users. However, the SIC decoding technology of NOMA may bring large delay, which needs to be solved when designing the NOMA based MAC protocol in the scenario of UAV assisted data collection.

\begin{figure}[!t]
	\centering
	\includegraphics[width=0.35\textwidth]{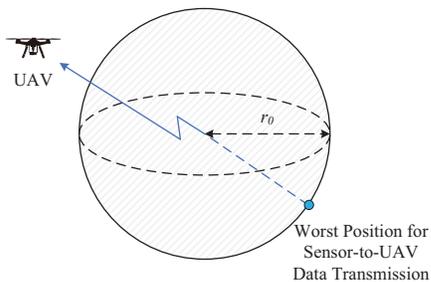}
	\caption{Data collection with uncertain position of sensor.}
	\label{fig_sensor_position}
\end{figure}

\subsection{Joint Sensing and Data Collection}

Sensors are usually randomly distributed in the target area. Most of the existing technologies that support positioning are not suitable for UAV assisted data collection. For example, the cost of installing GNSS chips in a large number of sensors is too high \cite{8030544}. And the energy consumption of GNSS positioning is high. Therefore, the position information of sensors is generally unknown before data collection. The joint sensing and data collection technology enables the UAV to locate sensors while data collection, as well as satisfying the requirements of low cost and low energy consumption of the IoT. 
	
	However, few literatures take sensing and data collection into consideration together in the scenario of UAV assisted data collection. In order to minimize the maximum energy consumption of all sensors, Wang \emph{et al.} \cite{8894454} apply a UAV as mobile anchor node and data collector to assist BSs in sensor positioning and data collection. A differential evolution algorithm is proposed to jointly optimize the trajectory of UAV and the transmission schedule of sensors over time. This paper provides a new scheme for the IoT, which realizes low-energy data collection and high-precision 3D positioning. At present, there are many separate studies of UAV assisted sensor positioning and UAV assisted data collection.

\subsubsection{UAV assisted sensor positioning}
Due to the flexibility of flight, UAV can select the appropriate position, adjust the posture, and locate the sensors in the target area from different angles, which effectively improves the positioning accuracy. In \cite{1}, Xu \emph{et al.} design a target location algorithm based on Unscented Kalman filter (UKF) to  improve the accuracy of passive target location of small UAV. The nonlinear target location system with additive white Gaussian noise (AWGN) is considered. The simulation results show that the mean square deviation of the filter estimation error has approached the Cramer-Rao lower bound of the nonlinear system. Han \emph{et al.} \cite{2010Development} propose an autonomous navigation system based on Microcontroller in \cite{1}, which obtains the 3D position of the target through the mathematical relationship between the UAV and the ground target. This method significantly improves the accuracy and confidence of measurement. Xu \emph{et al.} \cite{4} propose a positioning method combining maximum likelihood estimation (MLE) and square root volume Kalman filter (SRVKF). The MLE method is applied to locate the unknown sensors. The SRVKF algorithm is then introduced to further improve the positioning accuracy. In particular, the update strategy of threshold selection is used to reduce the influence of nonlinear factors. This method not only improves the positioning accuracy, but also greatly reduces the cost of loading GPS module. In the process of perceptual positioning, data loss or error is inevitable. Due to the uncertainty of wireless signal propagation, Li \emph{et al.} \cite{5} propose an adaptive positioning algorithm combining packet loss rate and received signal strength indicator (RSSI). It makes up for the measurement error without increasing the cost of wireless nodes. 

\subsubsection{UAV assisted data collection}
For UAV assisted data collection, most studies are carried out under the assumption that the sensor position is known. Generally, key performance indicators such as energy consumption of UAV and sensors, flight time of UAV and AoI will be taken as the objective functions of data collection algorithm. Zhan \emph{et al.} \cite{8119562} consider the general fading channel model between UAV and sensors and jointly optimize the sensors' wake-up schedule and trajectory of UAV with the goal of minimizing the maximum energy consumption of all sensors. In \cite{8698468}, You and Zhang initially optimize the communication scheduling of sensors and 3D trajectory of UAV in Rician Fading Channel to improve the data collection efficiency. Due to lacking the value of effective fading power, the parameters are approximated by logic function and solved by BCD and SCA. Han \emph{et al.} \cite{9321460} take AoI as an index to reveal the freshness of data, and analyze the single IoT device model and the multi-sensor model respectively with first come first service (FCFS) M/M/1 queuing model. In addition, Markov chain is used to predict the amount of collected data and packet loss rate, so as to collect data more efficiently.

In the future, low-power sensing and data collection need to be carried out simultaneously. Meanwhile, because the process of positioning is easily affected by the external environment \cite{3,2}, data collection algorithms with missing location information of sensors or positioning results with large errors need to be designed. Facing the positioning errors, the model of positioning uncertainty is considered in \cite{8894454}. The initially estimated position coordinates of sensors are taken as the sketchy prior information. Then, the uncertain space of each sensor position is modeled as a sphere with initially estimated position as the center of the sphere.  As shown in Fig. \ref{fig_sensor_position}, it is assumed that the sensor is located furthest from the UAV. B-Spline curves based UAV path is designed, which makes sensors complete the reliable data transmission under the worst data transmission conditions.

\section{Conclusion}

In this paper, we summarize the scenarios and related technologies of UAV assisted data collection for IoT. The key technologies are reviewed including clustering of SNs, UAV data collection mode, and joint path planning and resource allocation. Hovering mode, flying mode and hybrid mode of UAV are considered. In terms of joint path planning and resource allocation, according to the optimization algorithms used, we review the literatures from the perspectives of graph theory based algorithms, optimization theory based algorithms and AI based algorithms. Finally, we discuss the future trends from two aspects, i.e, efficient multiple access as well as joint sensing and data collection. This article may provide a reference for the research of UAV assisted data collection for IoT.

\section*{Acknowledgments}
The authors appreciate editor and anonymous reviewers for
their precious time and great effort in improving this paper.

\bibliographystyle{IEEEtran}

\bibliography{survey}

\end{document}